\documentclass{article}
\usepackage{microtype}
\usepackage{graphicx}
\usepackage{subcaption}
\usepackage{booktabs} 
\usepackage{multirow}
\usepackage{makecell}
\usepackage{array}
\usepackage{graphicx}
\usepackage{bm}
\usepackage{xspace}
\usepackage{hyperref}

\usepackage[accepted]{icml2026}
\usepackage{amsmath}
\usepackage{amssymb}
\usepackage{mathtools}
\usepackage{amsthm}
\usepackage{circledsteps}
\usepackage[capitalize,noabbrev]{cleveref}
\theoremstyle{plain}

\theoremstyle{definition}

\theoremstyle{remark}

\usepackage[table]{xcolor}
\definecolor{ourscolor}{HTML}{B2E6EA}

\newcommand{\name}{FourTune\xspace}
\usepackage[textsize=tiny]{todonotes}

\icmltitlerunning{\name{}: Towards Fully 4-Bit Efficient Post-Training for Diffusion Models}

\begin{document}
\defcitealias{schulman2025lora}{Schulman et al., 2025}

\makeatletter
\newcommand{\ForceICMLAffilNumber}[2]{%
  \@ifundefined{c@@affil#1}{\newcounter{@affil#1}}{}%
  \setcounter{@affil#1}{#2}%
}
\ForceICMLAffilNumber{nunchux}{1}
\ForceICMLAffilNumber{mit}{2}
\ForceICMLAffilNumber{cmu}{3}
\ForceICMLAffilNumber{stanford}{4}
\ForceICMLAffilNumber{ucb}{5}
\setcounter{@affiliationcounter}{5}
\makeatother

\newcommand{\icmlProjectLead}{\textsuperscript{\textdagger}Project lead }

\twocolumn[
  \icmltitle{\name: Towards Fully 4-Bit Efficient Post-Training for Diffusion Models}

  \icmlsetsymbol{equal}{*}
  \icmlsetsymbol{lead}{\textdagger}

  \begin{icmlauthorlist}
    \icmlauthor{Bowen Xue}{equal,stanford}
    \icmlauthor{Zihan Min}{equal,mit}
    \icmlauthor{Xingyang Li}{equal,mit}
    \icmlauthor{Zhekai Zhang}{nunchux}
    \icmlauthor{Haocheng Xi}{ucb}
    \icmlauthor{Lvmin Zhang}{stanford}
    \icmlauthor{Maneesh Agrawala}{stanford}
    \icmlauthor{Jun-Yan Zhu}{cmu}
    \icmlauthor{Song Han}{mit}
    \icmlauthor{Yujun Lin}{nunchux}
    \icmlauthor{Muyang Li}{nunchux}
  \end{icmlauthorlist}

  \icmlaffiliation{nunchux}{Nunchux AI}
  \icmlaffiliation{mit}{MIT}
  \icmlaffiliation{cmu}{CMU}
  \icmlaffiliation{stanford}{Stanford University}
  \icmlaffiliation{ucb}{UC Berkeley}

  \icmlcorrespondingauthor{Muyang Li}{\mbox{muyangli@nunchux.ai}}

  \icmlkeywords{}

  \vskip 0.3in

  \icmlkeywords{}
    {
    \begin{center}
        \vspace{-3mm}
        \centering
        \captionsetup{type=figure}
        \includegraphics[width=\textwidth]{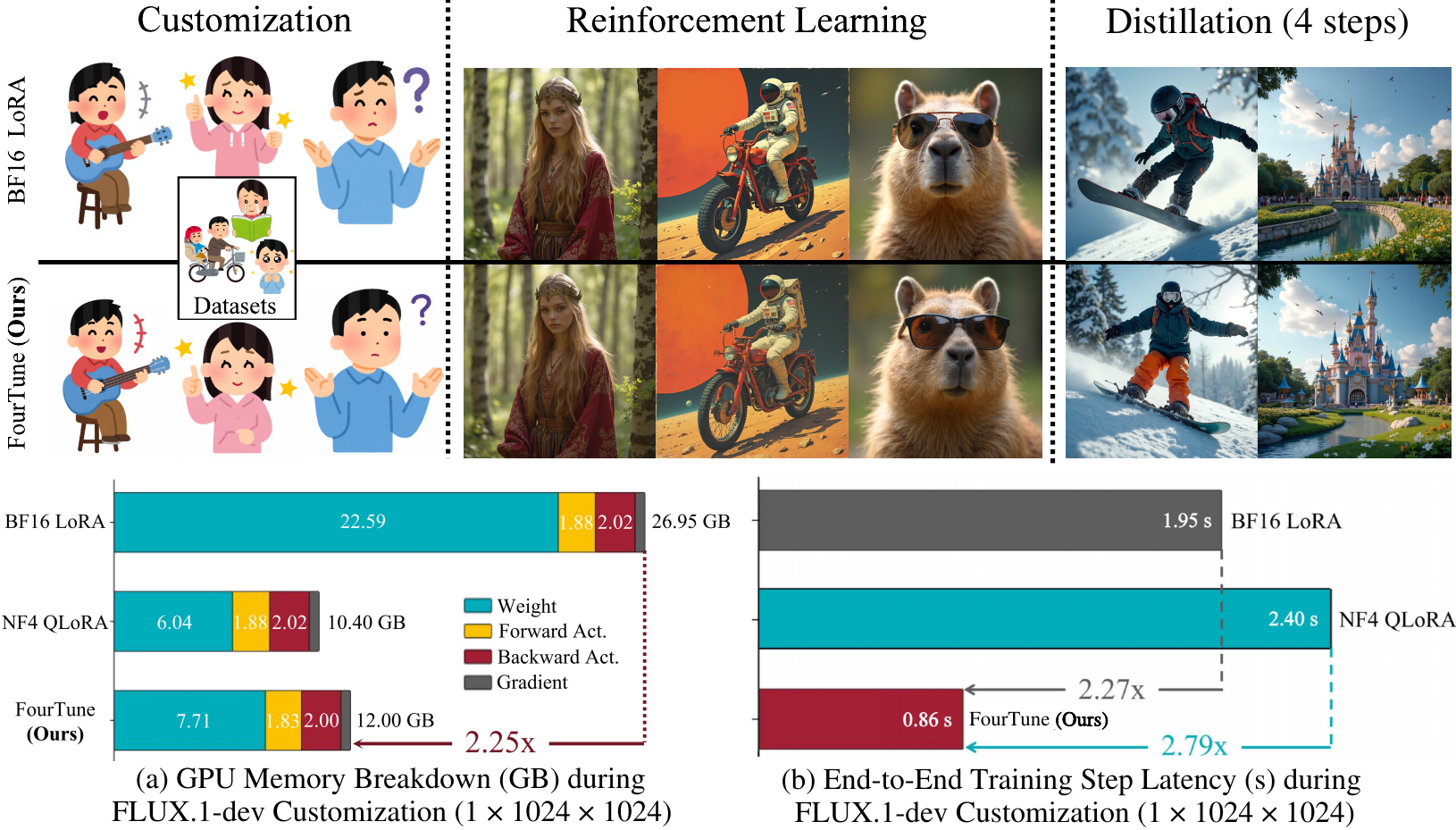}
        \caption{\textbf{Qualitative and quantitative comparison of \name against baselines.} 
        \textbf{Top:} Visual comparisons across three diverse post-training tasks: Customization, Reinforcement Learning, and Distillation. Despite extremely low-bit quantization (W4A4G4), \name produces high-fidelity images visually indistinguishable from the full-precision BF16 LoRA baseline.
        \textbf{Bottom:} Efficiency benchmarks performed on FLUX.1-dev ($1024\times1024$). 
        \textbf{(a)} \name reduces GPU memory consumption by \textbf{2.25$\times$} compared to BF16 LoRA, achieving a compact footprint comparable to NF4 QLoRA.
        \textbf{(b)} In terms of training speed, \name significantly outperforms existing methods, achieving \textbf{2.27$\times$} and \textbf{2.79$\times$} speedups over BF16 LoRA and NF4 QLoRA, respectively, effectively breaking the memory-speed trade-off in large model post-training.}
        \label{fig:teaser}
    \end{center}
    
    }
]
\printAffiliationsAndNotice{\icmlEqualContribution}



\begin{abstract}
Diffusion models have become a dominant paradigm for high-quality generative modeling, while post-training is essential for adapting them to diverse downstream applications.
However, post-training of large diffusion models is still challenging due to the prohibitive memory footprints and slow training speed, which existing parameter-efficient fine-tuning methods only partially address.
To overcome these limitations, we propose \textbf{\name}, an efficient post-training framework for diffusion models based on an end-to-end \textbf{W4A4G4} paradigm. \name introduces a triple-branch hybrid pipeline that augments the standard LoRA architecture with a frozen numerical stabilizer to isolate quantization-sensitive outliers, enabling stable training under native 4-bit computation. In addition, \name employs hardware-efficient block-wise quantization and customized fused kernels to support efficient quantized backpropagation and reduce memory bandwidth overhead.
Across customization, reinforcement learning, and distillation tasks, \name matches the quality of full-precision fine-tuning. On FLUX.1-dev (12B), \name reduces memory overhead by \textbf{2.25$\times$} and increases end-to-end training throughput by \textbf{2.27$\times$} compared to BF16 LoRA.

\end{abstract}
\section{Introduction}\label{sec:intro}

\begin{figure*}[t]
  \centering
  \includegraphics[width=\textwidth]{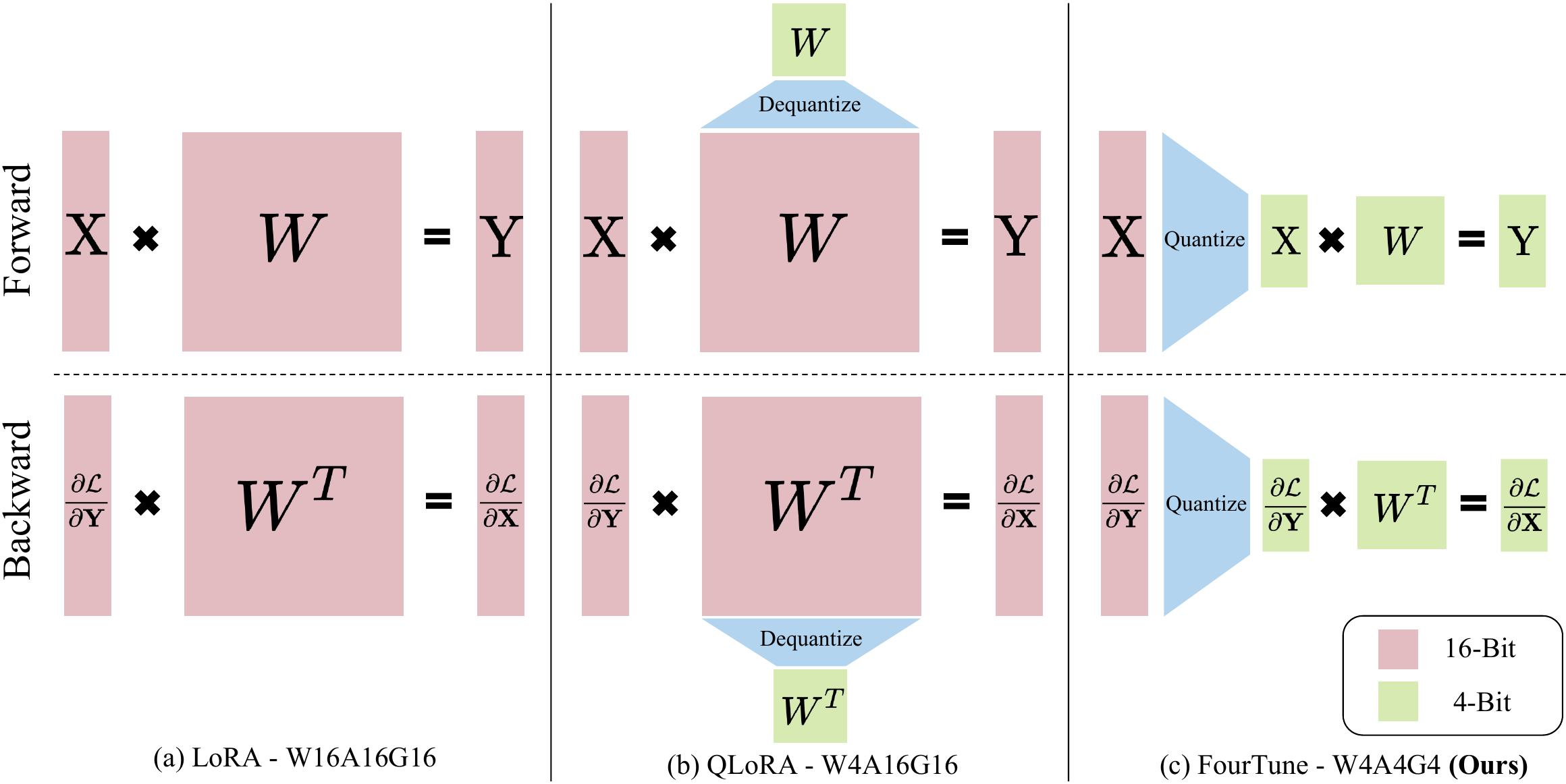}
  \caption{
Comparison of forward and backward pipelines across different PEFT methods.
\textbf{(a)} LoRA follows the standard high-precision training pipeline, where weights ($W$), activations ($A$), and gradients ($G$) are all stored and computed in high precision.
\textbf{(b)} QLoRA reduces the memory footprint by storing weights in 4-bit precision, but still requires on-the-fly dequantization to 16-bit for computation (W4A16G16).
\textbf{(c)} Our \name quantizes weights, activations, and gradients all to 4-bit (W4A4G4), enabling low-bit computation for both the forward and backward passes while preserving memory efficiency.}
  \label{fig:W4A4G4}
\end{figure*}

\looseness=-1
Generative models have demonstrated remarkable capabilities in synthesizing high-fidelity and semantically complex content~\cite{flux1,Qwen-Image}. Driven by the pursuit of higher generation quality, the community has scaled up model sizes to unlock greater potential, witnessing explosive parameter growth. Model sizes have increased from the 860M-parameter SD1.5~\cite{SD1.5}, to the 12B DiT-based FLUX.1~\cite{flux1}, and more recently to the 20B-parameter Qwen-Image~\cite{Qwen-Image}. However, this rapid scaling has substantially raised the computational requirements for both training and deployment, making it increasingly difficult to run such models on consumer-grade GPUs. Although advanced quantization methods~\cite{SVDQuant,Q-Diffusion,PTQ4DM} have lowered the barrier for inference via low-bit quantization, the resource bottleneck in post-training remains largely unaddressed.

Crucially, post-training serves as a vital pathway toward model practicality and personalization. Specifically, through customization, models can learn specific characters, styles, or concepts~\cite{custom-diffusion,DreamBooth}; via Reinforcement Learning, models can be aligned with human aesthetic preferences~\cite{SRPO,OnlineRL5}; and through distillation, the inference steps of diffusion models can be significantly reduced to save deployment costs~\cite{trajectory6,trajectory5}. However, the computation and memory requirements of the diffusion model post-training are still prohibitively high.
Consequently, achieving computationally efficient and memory-friendly post-training has emerged as a critical issue.

Recent parameter-efficient fine-tuning (PEFT) methods have gained increasing popularity by substantially reducing the number of trainable parameters and computational cost, while achieving performance comparable to full-parameter fine-tuning~\cite{schulman2025lora}.
However, despite these advances, existing PEFT approaches have not yet translated into further practical efficiency gains, as they remain fundamentally constrained by a persistent memory–speed trade-off.
For instance, LoRA~\cite{LoRA} (\Cref{fig:W4A4G4}(a)) reduces training overhead by minimizing the number of trainable parameters. Nevertheless, the full-precision backbone weights still occupy a substantial amount of GPU memory. To alleviate this bottleneck, QLoRA~\cite{QLoRA} (\Cref{fig:W4A4G4}(b)) reduces memory usage by quantizing the base model weights. However, it does not reduce computational cost, and the frequent online dequantization can even incur additional computational overhead.
In summary, a natural question arises: \textit{Is it possible to further accelerate post-training beyond LoRA?}

To address this challenge, we propose \textbf{\name}, a fully 4-bit post-training framework designed to break existing efficiency bottlenecks. Unlike methods like QLoRA that only quantize weights, \name adopts a fully quantized backbone, using 4-bit precision for weights, activations, and gradients (\textbf{W4A4G4}) (\Cref{fig:W4A4G4}(c)).
To the best of our knowledge, \name is the \textbf{first} fully 4-bit post-training framework for weight, activation, and gradient for large generative models.
We introduce a Numerical Stabilizer on top of the standard LoRA fine-tuning framework to ensure stable and convergent training under extremely low-bit quantization regimes.
We also employ block-wise quantization to enable efficient on-the-fly transposition for direct 4-bit backpropagation, eliminating transposition and dequantization overhead.
Moreover, we propose customized kernel fusion on the LoRA module and the MLP module to minimize memory access overhead.
This synergistic design reduces both computational load and memory footprint, achieving memory usage comparable to QLoRA while surpassing the training speed of full-precision LoRA fine-tuning.

\looseness=-1
We conduct extensive experiments across customization, reinforcement learning, and distillation tasks, showing that \textbf{\name} enables efficient post-training of diffusion models while matching the generation quality of full-precision fine-tuning. Compared to BF16 LoRA training, \name reduces memory footprint by up to 2.25$\times$ and accelerates training by up to 2.27$\times$. Overall, \name bridges the gap between high efficiency and high performance for diffusion-model post-training. 
\section{Related Work}\label{sec:related}
\subsection{Post-Training of Diffusion Models}
As diffusion models scale up~\cite{SDXL,flux1,Qwen-Image,Wan}, post-training has become essential for downstream adaptation of large diffusion models.
This phase typically includes customization, reinforcement learning, and distillation.

Customization enables models to learn new concepts from specific identities, styles, or subjects, while preserving the original generative fidelity~\cite{custom-diffusion, DreamBooth, TextualInversion}.
Reinforcement Learning incorporates human preferences into training through online RL~\cite{OnlineRL1,OnlineRL2,OnlineRL3}, which performs direct gradient updates on differentiable rewards, and policy-based approaches~\cite{policy1,policy2,policy3}.
Distillation compresses multi-step denoising into efficient few-step generation, using methods like trajectory regression~\cite{trajectory1,trajectory2,trajectory3}, consistency modeling~\cite{consistency1,consistency2,consistency3}, and distribution matching~\cite{distribution1,distribution2,distribution3}.

\subsection{Parameter-Efficient Adaptation}
Parameter-efficient fine-tuning (PEFT) has become a standard alternative to full fine-tuning, motivated by the observation that adaptation in large models often lies in a low intrinsic dimension~\cite{aghajanyan2020intrinsic}. Among PEFT methods, Low-Rank Adaptation (LoRA) and its variants~\cite{LoRA,T-LoRA,Ada-LoRA,DoRA} are particularly effective in both language models and diffusion models~(\citetalias{schulman2025lora}; \citealp{yeh2024navigating,trajectory6}), as they introduce low-rank update matrices while keeping pretrained weights frozen, thereby avoiding additional inference latency incurred by adapter-based methods~\cite{houlsby2019parameter}. 

To further improve fine-tuning efficiency, QLoRA~\cite{QLoRA} performs LoRA training with a 4-bit quantized backbone, substantially reducing memory consumption at the cost of increased computational overhead due to online dequantization. 
Building on this, subsequent works explore improved accuracy--efficiency trade-offs. For instance, LoftQ~\cite{LoftQ} tackles this through LoRA-aware quantization techniques, while QA-LoRA~\cite{QA-LoRA} reduces auxiliary overhead by integrating additional weight to the quantized base model.

\subsection{Model Quantization}
Quantization reduces memory and computation by mapping full-precision weights $\boldsymbol{W}$ to low-bit representations, typically via $\tilde{\boldsymbol{W}}=\mathrm{Round}(\boldsymbol{W}/s)$ with scaling factor $s$. Modern formats such as NF4 and hardware-aware floating-point schemes (e.g., MXFP4 and NVFP4) adopt group-wise scaling to better preserve numerical fidelity on contemporary accelerators~\cite{QLoRA,mxfp4,nvidia2024blackwell}. 

Prior work has shown that diffusion models can be effectively compressed to 8-bit or 4-bit precision for efficient inference~\cite{Q-Diffusion,PTQ4DM,SVDQuant}. More recent studies extend quantization to training, including pretraining large language models directly in 4-bit precision (NVFP4)~\cite{NVFP4}. Quantization has also been combined with parameter-efficient adaptation. For example, QeRL~\cite{QeRL} integrates NVFP4 with LoRA and demonstrates improved exploration in reinforcement learning settings.
\begin{figure*}[t]
    \includegraphics[width=\textwidth]{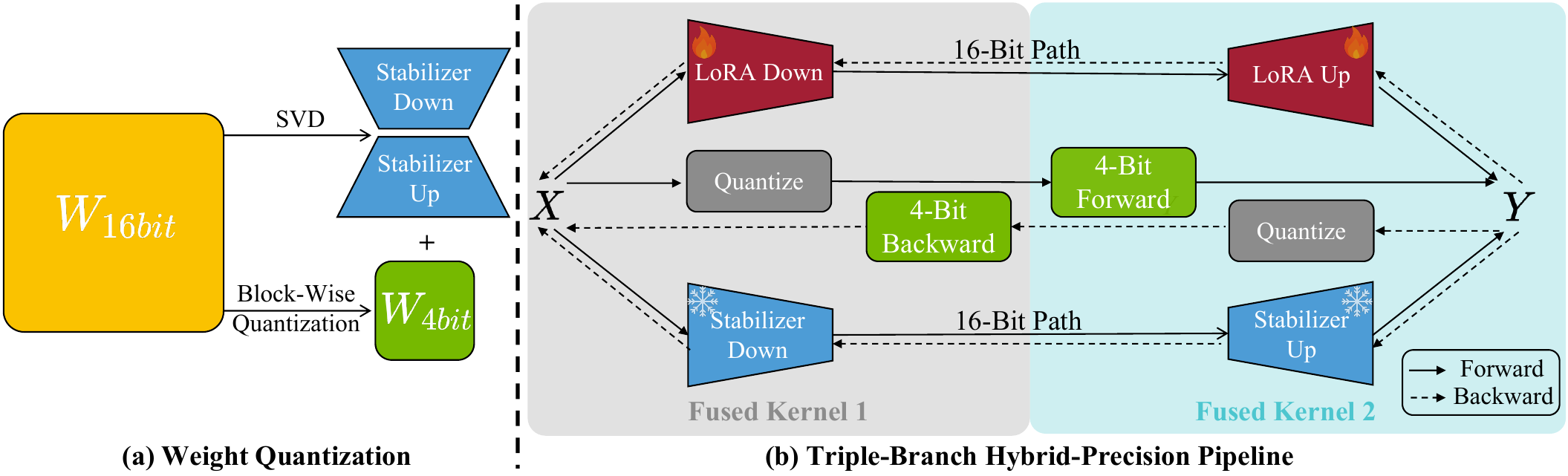}
    \caption{\textbf{Overview of the \name~framework.} (a) \textbf{Weight Quantization:} High-precision weights are decomposed via SVD into a quantization-friendly 4-bit residual and a high-precision stabilizer. (b) \textbf{Triple-Branch Pipeline:} The architecture consists of a frozen 4-bit backbone for arithmetic efficiency, a frozen stabilizer to ensure numerical precision, and a trainable LoRA branch for task adaptation.}
    \label{fig:pipeline}
\end{figure*}

\section{Method}\label{sec:method}

We propose \name, a post-training framework that establishes a native 4-bit pipeline for large generative models (\Cref{fig:pipeline}). Our method quantizes weights, activations, and gradients to 4-bit and executes backbone matrix multiplications directly in 4-bit. This design maximizes arithmetic efficiency while significantly reducing memory footprint. In this section, we describe the overall 4-bit training pipeline backed by a small full-precision stabilizer branch, a block-wise quantization strategy that enables efficient transposition during backpropagation, and customized kernel fusions that improve memory-bandwidth utilization.

\subsection{Triple-Branch Hybrid-Precision Pipeline}
Directly performing forward and backpropagation on aggressively quantized (e.g., 4-bit) weights is often numerically unstable, due to the limited dynamic range and the presence of high-magnitude outliers in weights, activations, and gradients. To address this challenge, we propose a \emph{triple-branch hybrid-precision pipeline} that explicitly decouples numerical stabilization from task adaptation.

Our method features an additional \emph{low-rank, full-precision stabilizer branch} to explicitly handle quantization-sensitive outliers and improve numerical stability during training. Concretely, after isolating the stabilizer component, the remaining frozen backbone weights are quantized using a hardware-friendly low-bit floating-point format (NVFP4) to leverage the powerful 4-bit computing units on GPUs.

Formally, building upon the spectral decomposition from SVDQuant~\cite{SVDQuant}, we decompose the pre-trained weight matrix $\boldsymbol{W} \in \mathbb{R}^{m \times n}$ into a quantization-friendly residual $R$ and a low-rank outlier component $\boldsymbol {L}_{\text{stab}}$:
\begin{equation}
    \boldsymbol{W} \approx \boldsymbol{R} + \boldsymbol{L}_{\text{stab}}, \quad \text{where } \boldsymbol{L}_{\text{stab}} = \boldsymbol{L_1 L_2}.
\end{equation}
Here, $\boldsymbol{L}_{\text{stab}}$ captures the high-magnitude outliers that are sensitive to quantization, while the residual exhibits a compressed value range suitable for 4-bit representation.

Distinct from previous approaches that utilize this decomposition solely for inference acceleration, we propose to utilize $\boldsymbol{L}_{\text{stab}}$ as a frozen ``numerical stabilizer'' during training.
This allows us to maintain the majority of parameters (the residual $\boldsymbol{R}$) in the 4-bit domain for computationally efficient matrix multiplication ($\text{GEMM}_{\text{4bit}}$), while isolating numerical risks in the high-precision branch.
To adapt the model to downstream tasks, we introduce a third, trainable low-rank branch $\boldsymbol{L}_{lora}$.
The forward propagation for an input $\boldsymbol{X}$ is thus formulated as a triple-branch computation:
\begin{equation}
\begin{aligned}
\boldsymbol{Y}
&= \text{GEMM}_{\text{4bit}}(Q(\boldsymbol{X}), Q(\boldsymbol{R}))
&& \text{\Circled{1} Frozen 4-Bit Backbone} \\
&\quad + \boldsymbol{X} \cdot \boldsymbol{L}_{\text{stab}}
&& \text{\Circled{2} Frozen Stabilizer} \\
&\quad + \boldsymbol{X} \cdot (\boldsymbol{AB})
&& \text{\Circled{3} Trainable Adapter}
\end{aligned}
\end{equation}
where $Q(\cdot)$ denotes the 4-bit quantization, applied offline for the residual weights $\boldsymbol{R}$ and dynamically for activations $\boldsymbol{X}$. In this pipeline, while error signals propagate through all branches to ensure correct gradient flow to preceding layers, weight gradients are computed exclusively for the adapter branch ($\boldsymbol{A}, \boldsymbol{B}$). Branches \Circled{1} and \Circled{2} function as frozen, differentiable pathways that facilitate backward propagation without requiring parameter updates.
This design enables native 4-bit computation for the backbone, significantly reducing memory bandwidth and computational latency.

\subsection{4-Bit Backward Pass via Block-Quantization}
In the backward pass, our pipeline minimizes computational redundancy by distinguishing between gradient propagation (required for all branches) and parameter updates (exclusive to the adapter). 
Let $\boldsymbol{G} = \frac{\partial \mathcal{{L}}}{\partial \boldsymbol{Y}}$ denote the incoming gradient.

Propagating gradients to the input requires computing $\frac{\partial \mathcal{L}}{\partial \boldsymbol{X}} = \boldsymbol{G W}^\top$. 
Leveraging the native 4-bit capability of our backbone, this operation is formulated as:
\begin{equation}
\begin{aligned}
\frac{\partial \mathcal{L}}{\partial \boldsymbol{X}} 
&= \text{GEMM}_{\text{4bit}}(Q(\boldsymbol{G}), Q(\boldsymbol{R})^\top)
&& \text{Backbone Pathway} \\
&\quad + \boldsymbol{G} \cdot \boldsymbol{L}_{\text{stab}}^\top
&& \text{Stabilizer Pathway} \\
&\quad + \boldsymbol{G} \cdot (\boldsymbol{AB})^\top
&& \text{Adapter Pathway}
\end{aligned}
\end{equation}
The critical challenge lies in the term $Q(\boldsymbol{R})^\top$, which requires performing matrix multiplication with a transposed 4-bit weight matrix. 
Standard per-output-channel quantization is ill-suited for this operation because the scaling factors, originally aligned with the output dimension, become aligned with the reduction dimension (accumulation axis) after transposition. 
This misalignment prevents the decoupling of scales from the inner product loop, making direct computation inefficient.

To overcome this, we employ a block-wise quantization strategy. 
By partitioning weights into small, independent blocks, quantization parameters are encapsulated locally. 
Crucially, this structure facilitates efficient online transposition: since scales are bound to local data blocks rather than global rows or columns, the computation kernel can fetch a 4-bit block and its corresponding scale simultaneously, regardless of whether the access pattern is transposed or non-transposed. 
This allows us to compute $Q(\boldsymbol{G})Q(\boldsymbol{R})^\top$ directly without dequantizing and re-quantizing the weights, maintaining the high throughput of 4-bit arithmetic.

Concurrently, we compute weight gradients exclusively for the trainable adapter branch. 
The gradients are derived using standard backpropagation chain rules:
\[
\frac{\partial \mathcal{L}}{\partial \boldsymbol{B}} = (\boldsymbol{X}\boldsymbol{A})^{\top}\boldsymbol{G},\quad 
\frac{\partial \mathcal{L}}{\partial \boldsymbol{A}} = \boldsymbol{X}^{\top}(\boldsymbol{G}\boldsymbol{B}^{\top}).
\]
\subsection{Kernel Fusion for Efficiency}

\paragraph{LoRA Fusion.}
While the stabilizer branch and the trainable low-rank branch introduce negligible additional computation, executing them as independent pathways can incur significant latency due to increased memory access overhead. In this regime, the system bottleneck shifts from computation to memory bandwidth. To mitigate this, we leverage the fact that all three branches share the same input during the quantization and down-projection phases, as illustrated in \Cref{fig:pipeline}(b). Accordingly, we fuse these operations into a single kernel. Furthermore, since the outputs of these branches must be accumulated, we fuse the 4-bit backbone computation with the up-projection operations to eliminate redundant memory traffic. This strategy substantially reduces the inference latency overhead typically introduced by the LoRA architecture.
\paragraph{MLP Fusion.}
In MLP blocks, intermediate activation tensors often possess high dimensionality (e.g., $3072 \times 12288$). Applying activation functions such as GeLU on these tensors becomes memory-bound when data is repeatedly written to and read from global memory. To address this, we fuse the first fully connected (FC1) GEMM with the subsequent FC2 quantization kernel. This approach allows intermediate results to be consumed immediately without materializing large tensors in global memory, thereby optimizing bandwidth utilization and accelerating end-to-end execution.

\renewcommand{\arraystretch}{1.4}
\begin{table*}[t]
    \centering
    \caption{Quantitative results on Customization tasks.}
    \label{tab:customization}
    \resizebox{\textwidth}{!}{%
        \begin{tabular}{ccccccccccc}
            \toprule
            \multirow{2.5}{*}{\textbf{Task}} & \multirow{2.5}{*}{\textbf{Method}} & \multirow{2.5}{*}{\textbf{Precision}} & 
            \multicolumn{4}{c}{\textbf{FLUX.1-dev}} & \multicolumn{4}{c}{\textbf{Qwen-Image}} \\
            \cmidrule(lr){4-7} \cmidrule(lr){8-11}
             & & & Similarity & \makecell{Image \\ Quality} & Diversity & \makecell{Prompt \\ Following} & 
            Similarity & \makecell{Image \\ Quality} & Diversity & \makecell{Prompt \\ Following} \\
            \midrule
            
            \multirow{4}{*}{\textbf{Identity}} 
            & BF16 LoRA & W16A16G16 & 0.771 & \underline{33.49} & \textbf{0.577} & \textbf{0.941} & \underline{0.601} & \underline{24.54} & \underline{0.742} & \textbf{0.998} \\
            & FP8 LoRA & W8A8G8    & \textbf{0.783} & 30.70 & 0.532 & 0.912 & 0.597 & 21.66 & \textbf{0.767} & \textbf{0.998} \\
            & NF4 QLoRA& W4A16G16  & \underline{0.780} & 32.27 & 0.530 & \underline{0.929} & 0.597  & \textbf{25.63} & 0.685 & 0.994 \\
            \rowcolor{ourscolor!40} \cellcolor{white} & \textbf{Ours} & W4A4G4    & \textbf{0.783} & \textbf{34.77} & \underline{0.570} & 0.913 & \textbf{0.612} & 24.40 & 0.695 & \underline{0.995} \\
            \midrule
            
            \multirow{4}{*}{\textbf{Style}} 
            & BF16 LoRA & W16A16G16 & 0.806 & 32.02 & 0.520 & 0.864 & 0.696 & 27.34 & \underline{0.683} & \textbf{0.999} \\
            & FP8 LoRA & W8A8G8    & \textbf{0.821} & 30.05 & \textbf{0.605} & \textbf{0.893} & 0.692 & 27.29 & 0.676 & \textbf{0.999} \\
            & NF4 QLoRA& W4A16G16  & 0.806 & \underline{32.19} & 0.529 & \underline{0.883} & \textbf{0.710} & \textbf{33.72} & 0.646 & \underline{0.928} \\
            \rowcolor{ourscolor!40} \cellcolor{white} & \textbf{Ours} & W4A4G4    & \underline{0.812} & \textbf{34.87} & \underline{0.547} & 0.874 & \underline{0.701} & \underline{29.88} & \textbf{0.684} & \textbf{0.999} \\
            \midrule
            
            \multirow{4}{*}{\textbf{Subject}} 
            & BF16 LoRA & W16A16G16 & \underline{0.731} & \textbf{16.57} & 0.415 & \textbf{0.946} & 0.618 & 16.92 & \underline{0.707} & \textbf{0.999} \\
            & FP8 LoRA & W8A8G8    & \textbf{0.733} & 15.72 & \textbf{0.460} & \underline{0.943} & 0.558 & \underline{18.02} & \textbf{0.714} & \textbf{0.999} \\
            & NF4 QLoRA & W4A16G16  & 0.724 & 15.30 & \underline{0.418} & 0.896 & \underline{0.695} & 15.06 & 0.639 & \underline{0.992} \\
            \rowcolor{ourscolor!40} \cellcolor{white} & \textbf{Ours} & W4A4G4    & \textbf{0.733} & \underline{16.22} & 0.415 & 0.920 & \textbf{0.703} & \textbf{26.92} & 0.689 & 0.983 \\
            
            \bottomrule
        \end{tabular}%
    }
\end{table*}
\section{Experiments}\label{sec:exp}
To evaluate the performance and versatility of \textbf{\name} across different computational tiers, we conduct all experiments on platforms powered by the NVIDIA Blackwell architecture, including 8$\times$ NVIDIA RTX Pro 6000 GPUs and 8$\times$ NVIDIA RTX 5090 consumer GPUs. Leveraging Blackwell's 4-bit Tensor Cores, we enable a realistic evaluation of both training and inference efficiency.

\subsection{Experimental Settings}
We evaluate \name across three core post-training paradigms: Customization, Reinforcement Learning, and Distillation. Please refer to our appendix for more details.

\paragraph{Customization Setup.}
We evaluate \name on a hybrid customization benchmark built from Custom Diffusion~\cite{custom-diffusion} and web-collected data, covering three tasks: Human Identity, Artistic Style, and General Subject. Experiments are conducted on FLUX.1-dev (12B)~\cite{flux1} and Qwen-Image (20B)~\cite{Qwen-Image} using LoRA with rank $r=64$ and identical hyperparameters. Evaluation uses task-specific feature metrics: AntelopeV2 for Human Identity~\cite{insightface_pypi,deng2019arcface}, CLIP for Artistic Style~\cite{clip}, and DINOv3 for General Subject~\cite{DINOv3}, with PyIQA for image quality~\cite{pyiqa} and BLIP for prompt following~\cite{BLIP}.

\begin{figure}[h]
    \centering
    \includegraphics[width=\linewidth]{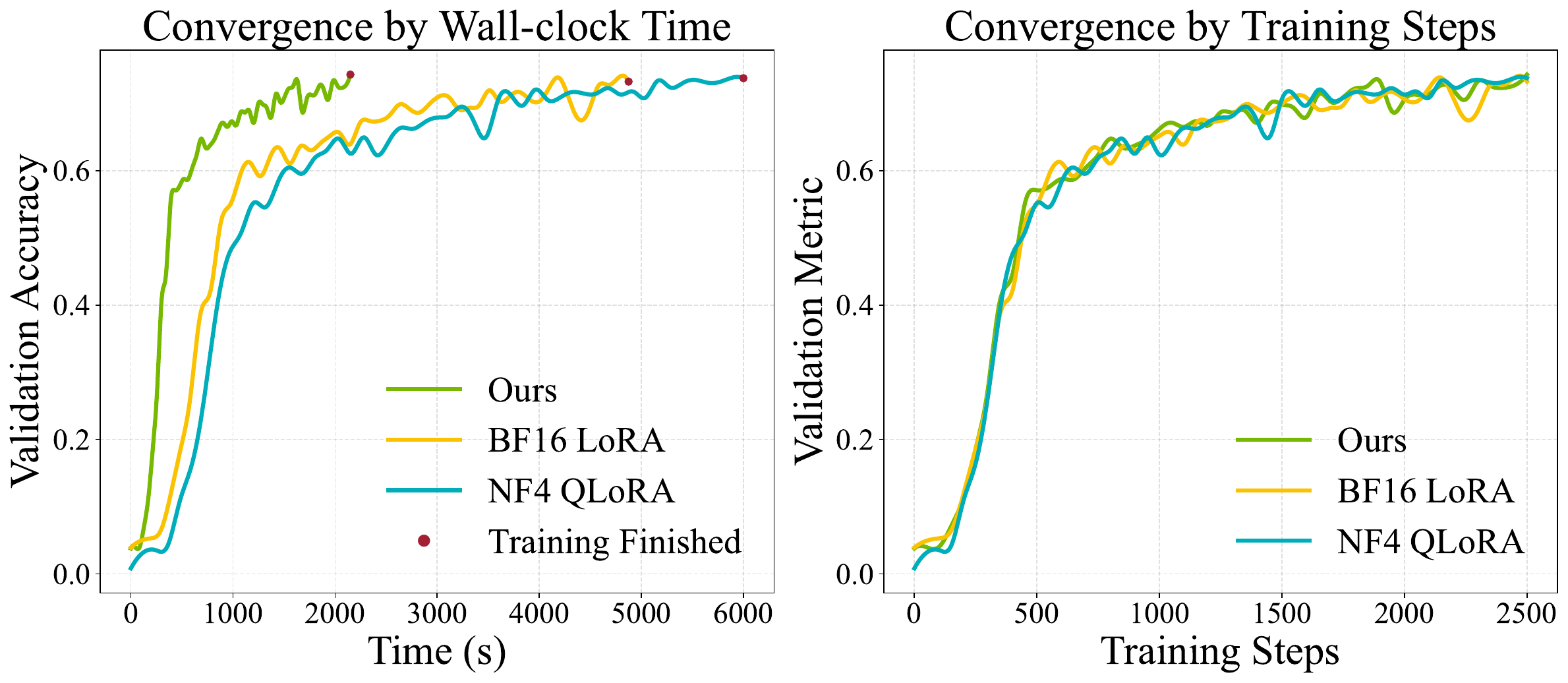}
    \caption{Validation curves for customization tasks, showing performance on par with full-precision training and with substantial acceleration.}
    \label{fig:curve}
\end{figure}

\begin{table*}[t]
\centering
\small
\setlength{\tabcolsep}{6pt}
\caption{Quantitative results for model distillation using the $\pi$-Flow algorithm. We evaluate the 4-step student model on HPSv2 and COCO-10k benchmarks. Performance is measured in terms of teacher alignment (FID), prompt alignment (CLIP), and preference alignment (HPSv2.1). Our W4A4G4 method maintains generation quality comparable to the 16-bit LoRA baseline.}
\label{tab:distill_eval}
\begin{tabular}{cccccccc}
\toprule
\multirow[c]{3}{*}[-1.5ex]{Method} & \multirow[c]{3}{*}[-1.5ex]{Precision} &
\multicolumn{3}{c}{HPSv2 prompts} &
\multicolumn{3}{c}{COCO-10k prompts} \\
\cmidrule(lr){3-5}\cmidrule(lr){6-8}
& & \multicolumn{1}{c}{Teacher align.} & \multicolumn{1}{c}{Prompt align.} & \multicolumn{1}{c}{Pref. align.}
& \multicolumn{1}{c}{Teacher align.} & \multicolumn{1}{c}{Prompt align.} & \multicolumn{1}{c}{Pref. align.} \\
\cmidrule(lr){3-3}\cmidrule(lr){4-4}\cmidrule(lr){5-5}
\cmidrule(lr){6-6}\cmidrule(lr){7-7}\cmidrule(lr){8-8}
& & FID$\downarrow$ & CLIP$\uparrow$ & HPSv2.1$\uparrow$
& FID$\downarrow$ & CLIP$\uparrow$ & HPSv2.1$\uparrow$ \\
\midrule
\textbf{BF16 LoRA}  & W16A16G16 & \textbf{15.50} & \textbf{0.288} & \underline{0.316} & \textbf{5.90}  & \textbf{0.269} & \textbf{0.311} \\
\textbf{SparseLoRA} & W16A16G16 & 202.51 & 0.261 & 0.245 & 75.45 & 0.264 & 0.238 \\
\textbf{NF4 QLoRA} & W4A16G16  & \underline{15.51} & \underline{0.287} & 0.310 & \underline{6.30} & \textbf{0.269} & 0.307 \\
\rowcolor{ourscolor!40} \textbf{Ours}  & W4A4G4    & \textbf{15.50} & 0.283 & \textbf{0.317} & 6.70  & \underline{0.266} & \underline{0.310} \\
\bottomrule
\end{tabular}
\end{table*}
\paragraph{Reinforcement Learning Setup.}
We evaluate \name in a reinforcement learning setting using SRPO~\cite{SRPO}, applying preference gradients only to LoRA adapters for parameter-efficient optimization. Experiments follow the original SRPO protocol with FLUX.1-dev as the backbone, trained on HPDv2 and guided by the HPSv2.1 reward model~\cite{HPDv2}.
Evaluation is performed using Aesthetic Score v2.5~\cite{discus0434_aestheticpredictor_v25}, PickScore~\cite{PickScore}, ImageReward~\cite{xu2023imagereward}, and SGP-HPS~\cite{SRPO}.

\paragraph{Distillation Setup.}
We evaluate \name on model distillation using $\pi$-Flow~\cite{trajectory6}, comparing against BF16 LoRA, NF4 QLoRA, and SparseLoRA~\cite{SparseLoRA}. Following the original $\pi$-Flow protocol, we distill FLUX.1-dev~\cite{flux1} into a 4-NFE student in a data-free setting. Evaluation is conducted on COCO-10k and HPSv2 prompts~\cite{HPDv2}, reporting Teacher-FID for quality, CLIP~\cite{clip} for prompt alignment, and HPSv2.1 for preference alignment.

\subsection{Main Results}
\begin{table}[h]
\centering
\small
\caption{Quantitative results on Reinforcement Learning tasks using the SRPO algorithm. }
\label{tab:rl_results}
\begin{tabular}{ccccc}
\toprule
Method & Aes & PickScore & IR & SGP-HPS \\
\midrule
\textbf{Base model} & 6.0135 & 0.2300 & \textbf{1.1189} &  0.0015 \\
\textbf{BF16 Finetune}  & \textbf{6.3447} & \underline{0.2316} & 1.0209 & \textbf{0.0029} \\
\textbf{BF16 LoRA}      & 6.1832 & 0.2307 & 1.0735 & 0.0020 \\
\textbf{NF4 QLoRA}     & 6.2189 & \textbf{0.2318} & \underline{1.0931} & 0.0020 \\
\rowcolor{ourscolor!40} \textbf{Ours}      & \underline{6.3119} & 0.2308 & 1.0152 & \underline{0.0027} \\
\bottomrule
\end{tabular}
\end{table}

\looseness=-1
\paragraph{Accuracy Results. }
\begin{figure}[h]
    \centering
    \includegraphics[width=\linewidth]{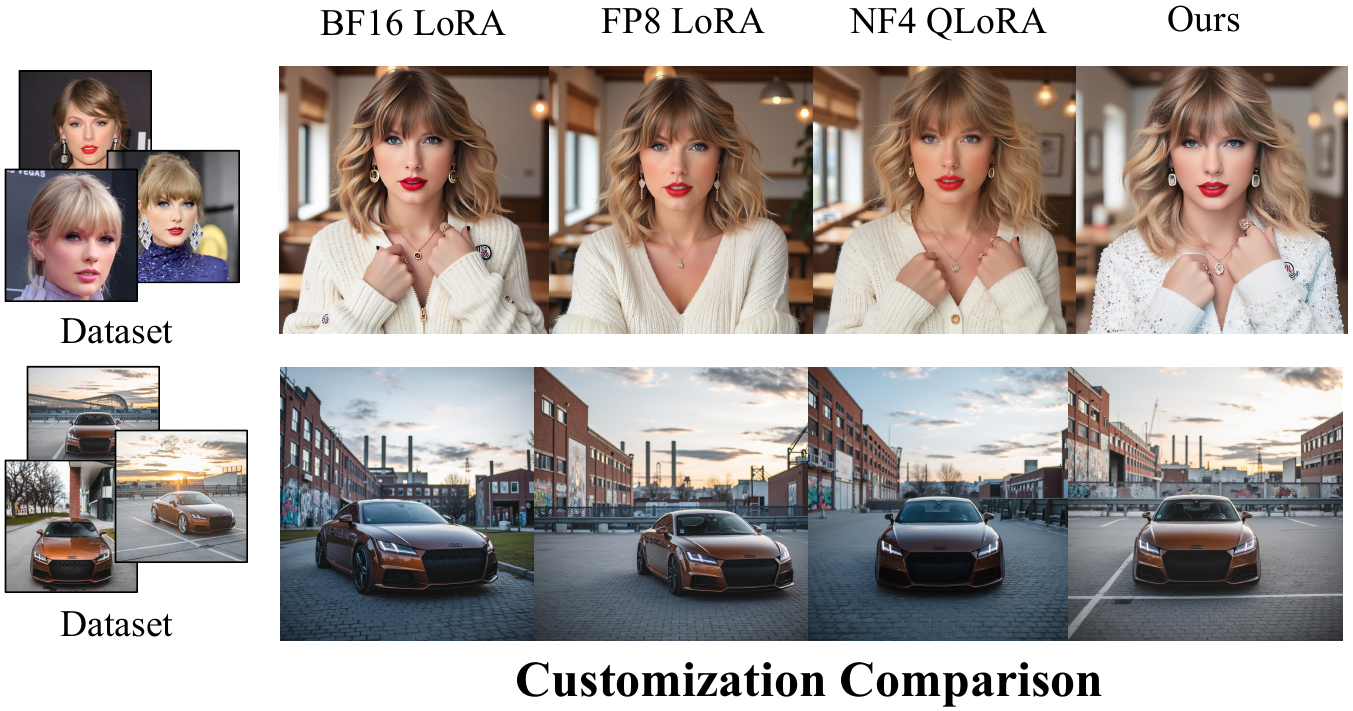}
    \caption{Qualitative comparison on customization. \name matches the generation quality of full-precision LoRA.}
    \label{fig:customization}
\end{figure}

For \textbf{Customization} (Table~\ref{tab:customization} and Figure~\ref{fig:customization}), \name scales post-training to models with up to 20B parameters. It matches the generation quality of full-precision LoRA across human identity preservation, artistic style transfer, and general subject reconstruction, validating its ability to capture fine-grained visual details.

\begin{figure}[h]
    \centering
    \includegraphics[width=\linewidth]{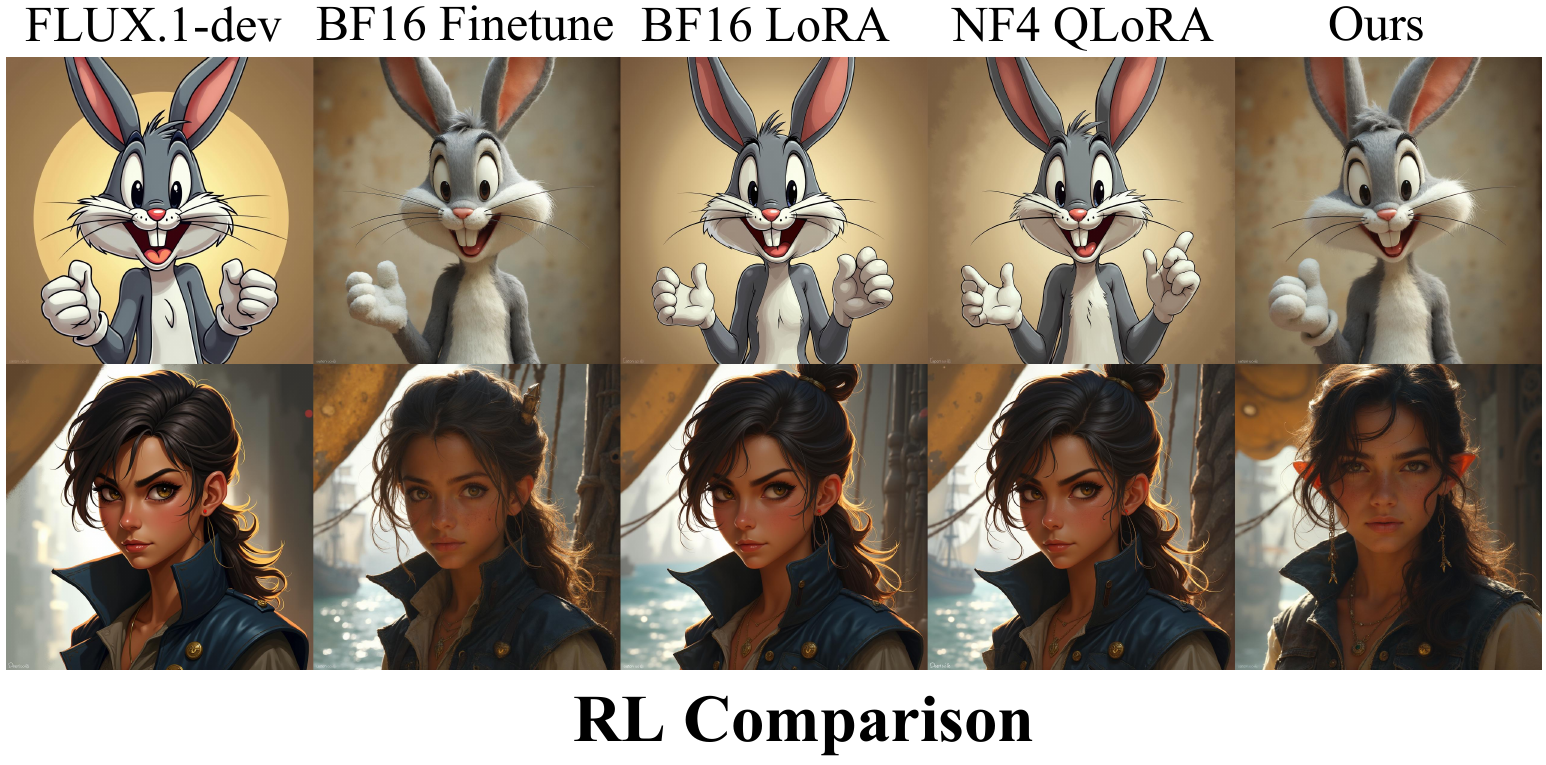}
    \caption{Qualitative comparison on RL. \name even matches the performance of full-precision, full-parameter finetuning.}
    \label{fig:RL}
\end{figure}

For \textbf{Reinforcement Learning} (Table~\ref{tab:rl_results} and Figure~\ref{fig:RL}), \name achieves performance comparable to even the full-precision, full-parameter fine-tuning. This indicates that the 4-bit training pipeline retains numerical fidelity to support the fine-grained updates in RL training.

\begin{figure}[h]
    \centering
    \includegraphics[width=\linewidth]{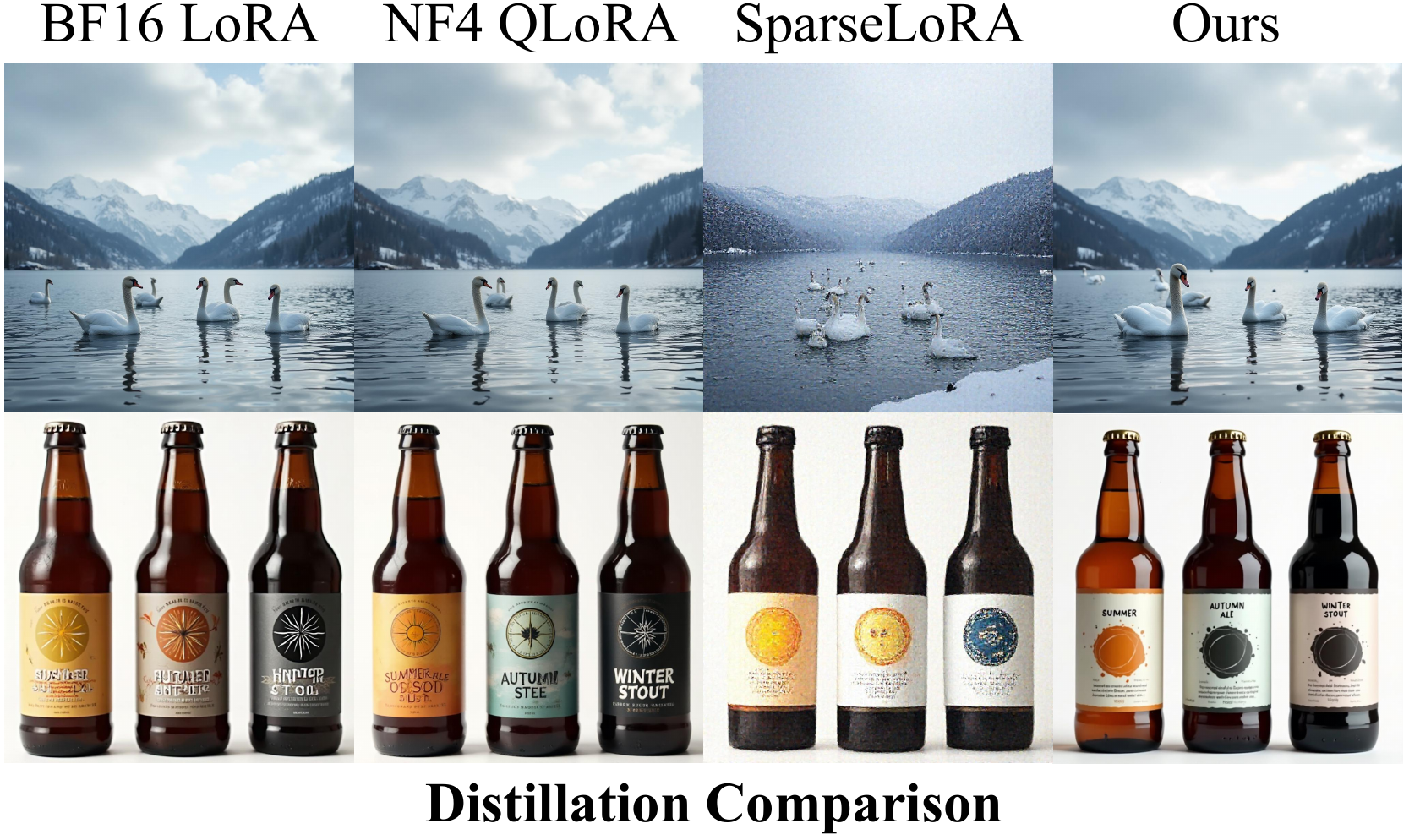}
    \caption{Qualitative comparison on distillation. \name closely matches the visual quality of BF16 LoRA and NF4 QLoRA, while SparseLoRA exhibits noticeable artifacts.}
    \label{fig:ditillation}
\end{figure}

For \textbf{Distillation} (Table~\ref{tab:distill_eval} and Figure~\ref{fig:ditillation}), \name can match the quality of the original 16-bit $\pi$-Flow implementation. Meanwhile, SparseLoRA underperforms \name even at 50\% sparsity, likely because timestep distillation is highly accuracy-sensitive, whereas SparseLoRA imposes only coarse-grained column sparsity.
These results suggest that \name remains stable even for distillation, improving efficiency without compromising generation quality.

\paragraph{Efficiency Results. }
We first evaluate the efficiency of \name on the DiT component of FLUX.1-dev.
By quantizing weights to 4-bit, the weight memory consumption is 7.71 GB, compared to 22.59 GB for BF16 LoRA,
achieving a \textbf{2.93$\times$} reduction.
For a single DiT training step, \name achieves a total latency of 612.4 ms
whereas BF16 LoRA requires 1541.0 ms,
corresponding to a \textbf{2.52$\times$} speedup.

We further measure end-to-end training efficiency on the customization task.
\name reduces the overall memory consumption to 14.51 GB, compared to 29.35 GB required by BF16 LoRA, corresponding to a \textbf{2.02$\times$} reduction.
On NVIDIA RTX 5090, \name achieves a single-step training time of 0.86 s, corresponding to a \textbf{2.27$\times$} speedup over BF16 LoRA and a \textbf{2.79$\times$} speedup over NF4 QLoRA.
On NVIDIA RTX Pro 6000, \name achieves a single-step training time of 0.66 s, yielding a \textbf{2.18$\times$} speedup compared to BF16 LoRA and a \textbf{2.85$\times$} speedup over NF4 QLoRA.

These results demonstrate that \name
not only reduces memory consumption substantially,
but also consistently delivers practical end-to-end training acceleration.

\subsection{Ablation Study}
To rigorously evaluate the effectiveness of our proposed method, we conducted ablation studies focusing on four key aspects: training precision configurations, the impact of the Stabilizer, the impact of our block-level quantization, and the efficiency gains from kernel fusion optimizations.
\paragraph{Impact of Precision Configurations.}
We introduce a W4A4G4 training framework that achieves performance comparable to full-precision training. To investigate how the precision of different components affects final training outcomes, we conducted experiments across four different precision settings (W4A4G4, W4A16G4, W4A4G16, and W4A16G16) on RL tasks. Specifically, we performed complete training on a subset of 500 prompts from the HPDv2 Benchmark. It is important to note that among these configurations, only W4A4G4 fully benefits from the acceleration provided by 4-bit computation.

As demonstrated in Table \ref{tab:ablation_precision}, variations in component precision do not lead to performance degradation. Notably, our W4A4G4 configuration enables the model to leverage significant hardware acceleration while maintaining performance alignment with full-precision baselines.
\begin{table}[h]
\centering
\small
\caption{Ablation study on different precision configurations. Note that W4A4G4 is the only setting that fully utilizes 4-bit acceleration.}
\label{tab:ablation_precision}
\small
\begin{tabular}{ccccc}
\toprule
Method & Aes & PickScore & IR & SGP-HPS \\
\midrule
\textbf{W4A16G16}  & 6.0816 & 0.2352 & \textbf{1.2231} & 0.0002 \\
\textbf{W4A4G16}      & 6.0308 & \textbf{0.2357} & 1.1982 & \underline{0.0005} \\
\textbf{W4A16G4}     & \underline{6.1386} & \underline{0.2354} & \underline{1.2196} & 0.0001 \\
\rowcolor{ourscolor!40} \textbf{W4A4G4 (Ours)}      & \textbf{6.1779} & 0.2353 & 1.1712 & \textbf{0.0008}\\
\bottomrule
\end{tabular}
\end{table}

\begin{figure}[h]
    \centering
    \includegraphics[width=\linewidth]{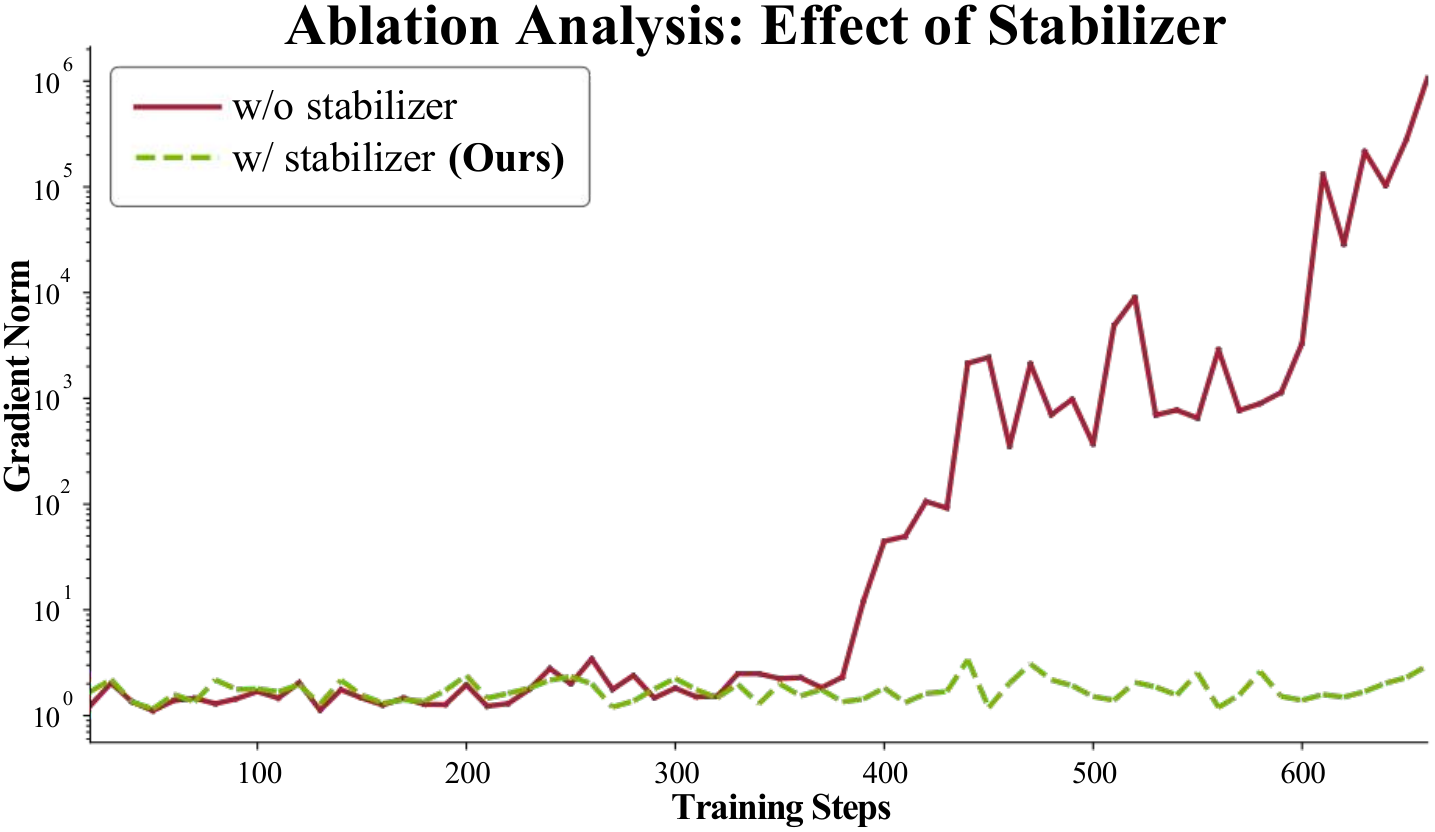}
    \caption{\textbf{Effect of the Stabilizer on numerical stability.} Comparison of gradient norms with and without the proposed Stabilizer. The results show that our method prevents the gradient explosion observed in the baseline setting (red line), ensuring stable convergence.}
    \label{fig:stab}
\end{figure}

\paragraph{Effectiveness of the Stabilizer.}
During native 4-bit training, the accumulation of quantization errors occasionally triggers gradient explosion, rendering the training process ineffective. To investigate this, we monitored the gradient norms over training steps in knowledge distillation experiments, comparing settings with and without the Stabilizer. As illustrated in Figure~\ref{fig:stab}, the naive W4A4G4 pipeline suffers from gradient explosion, while our \name's gradient norms remain stable. These results demonstrate that our Stabilizer effectively maintains numerical stability, thereby preventing gradient explosion and ensuring consistent convergence.

\paragraph{Impact of Quantization Granularity.}
As detailed in Section~\ref{sec:method}, we adopt block-wise quantization with a $16 \times 16$ granularity to resolve the non-contiguous memory access issues inherent to standard group-wise quantization ($1 \times 16$) during backpropagation.
To verify that this coarser granularity does not compromise representation accuracy, we explicitly compare the performance of block-wise versus group-wise strategies on FLUX.1-dev. We measure reconstruction fidelity on 5,000 samples from the MJHQ-30K dataset against the BF16 baseline.
As presented in Table~\ref{tab:quant_ablation}, our block-wise strategy yields LPIPS and PSNR metrics comparable to the fine-grained group-wise approach. These results confirm that our design secures significant efficiency gains in the backward pass without incurring any noticeable degradation in generation quality.

\begin{figure}[h]
    \centering
    \includegraphics[width=\linewidth]{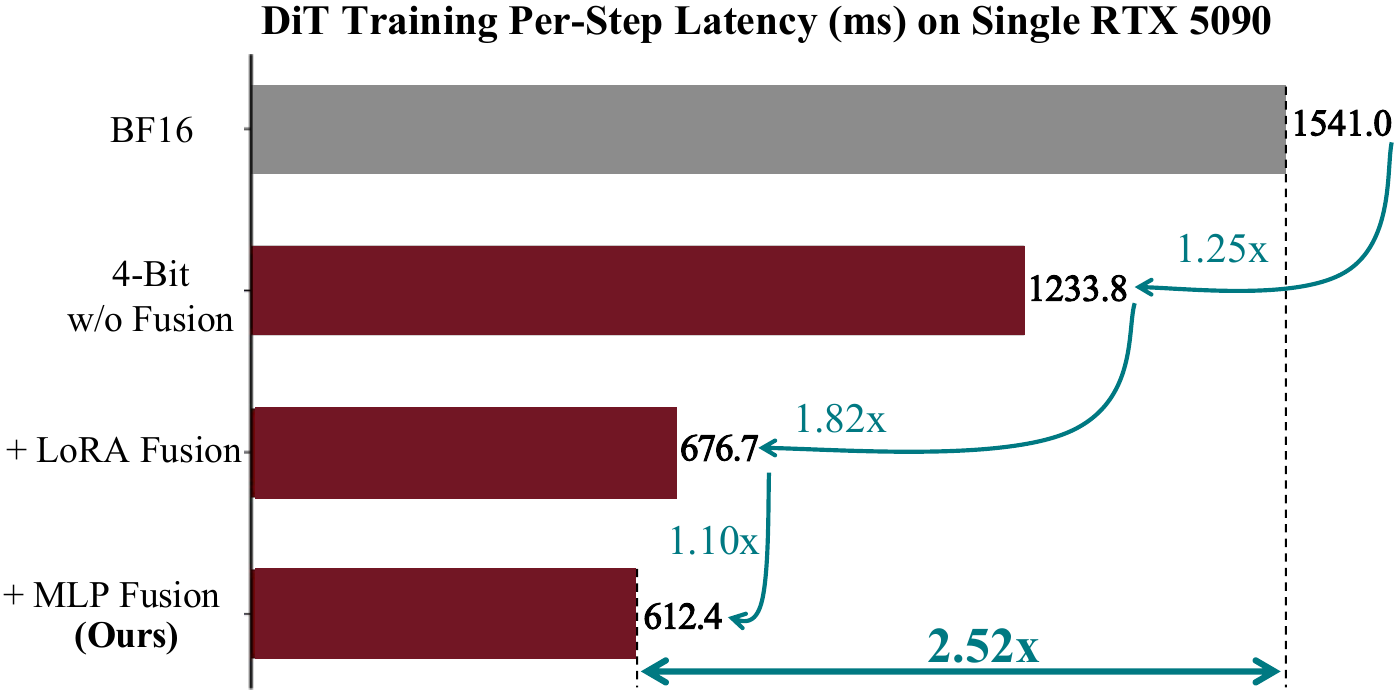}
    \caption{Breakdown of efficiency gains from different kernel fusion strategies. The combined optimization results in a 2.52$\times$ speedup relative to 16-bit LoRA fine-tuning.}
    \label{fig:kernel_breakdown}
\end{figure}

\begin{table}[h]
\centering
\small
\setlength{\tabcolsep}{8pt}
\caption{Quantitative comparison of quantization strategies on FLUX.1-dev. We evaluate reconstruction quality on 5,000 samples from the MJHQ-30K dataset compared to the BF16 baseline. Block-wise quantization ($16 \times 16$) maintains high fidelity comparable to the finer-grained group-wise ($1 \times 16$) strategy.}
\label{tab:quant_ablation}
\small
\begin{tabular}{cccc}
\toprule
Quantization Strategy & Granularity & LPIPS $\downarrow$ & PSNR $\uparrow$ \\
\midrule
\textbf{group-wise} & $1 \times 16$ & \textbf{0.203} & \textbf{21.5} \\
\rowcolor{ourscolor!40} \textbf{block-wise (Ours)} & $16 \times 16$ & 0.227 & 20.4 \\
\bottomrule
\end{tabular}
\end{table}
\paragraph{Kernel Fusion Breakdown.}
To analyze the sources of efficiency gains from Kernel Fusion, we decompose the contributions of individual fusion strategies, as shown in Figure~\ref{fig:kernel_breakdown}. Fusing LoRA-related kernels yields a 1.82$\times$ speedup over the unfused 4-bit baseline, while further fusing the MLP components provides an additional 1.10$\times$ acceleration. When all kernels are jointly fused, the combined effect results in an overall 2.52$\times$ speedup compared to standard LoRA, empirically demonstrating the effectiveness and complementarity of kernel fusion.

\section{Conclusion}\label{sec:conclu}
We have presented \name, a fully 4-bit post-training framework that enables end-to-end W4A4G4 optimization for large diffusion models by using a frozen stabilizer branch, block-wise quantization for efficient backward computation, and low-bit fused kernels. Extensive experiments across customization, reinforcement learning, and distillation demonstrate that \name matches full-precision baselines while substantially reducing memory footprint and improving throughput. This work contributes to efficient post-training of diffusion models.

\nocite{langley00}

\newpage
\section*{Impact Statement}
This work improves the efficiency of post-training large diffusion models through native 4-bit optimization, reducing computational and energy costs and making model adaptation more accessible.
However, easier adaptation may also lower the barrier to misuse, such as generating deceptive or harmful content.
In our release, we follow the licenses of the base models and datasets, and we encourage downstream users to apply appropriate safety filters and responsible deployment practices.



\section*{Acknowledgements}

We thank MIT-IBM Watson AI Lab, MIT and Amazon Science Hub, MIT AI Hardware Program,
National Science Foundation, Packard Foundation, Dell, LG, Hyundai, and Samsung for
supporting this research. We thank NVIDIA for donating the DGX server.

\bibliography{reference}

@String(CVPR  = {Conference on Computer Vision and Pattern Recognition})

@String(ICCV  = {International Conference on Computer Vision})

@String(NeurIPS  = {Conference and Workshop on Neural Information Processing Systems})

@String(ICLR  = {International Conference on Learning Representations})

@String(ICML = {International Conference on Machine Learning})

@String(ECCV  = {European Conference on Computer Vision})

@inproceedings{SD1.5,
  author       = {Robin Rombach and
                  Andreas Blattmann and
                  Dominik Lorenz and
                  Patrick Esser and
                  Bj{\"{o}}rn Ommer},
  title        = {High-Resolution Image Synthesis with Latent Diffusion Models},
  booktitle    = CVPR,
  year         = {2022},
}

@inproceedings{SDXL,
  author       = {Dustin Podell and
                  Zion English and
                  Kyle Lacey and
                  Andreas Blattmann and
                  Tim Dockhorn and
                  Jonas M{\"{u}}ller and
                  Joe Penna and
                  Robin Rombach},
  title        = {{SDXL:} Improving Latent Diffusion Models for High-Resolution Image Synthesis},
  booktitle    = ICLR,
  year         = {2024},
}

@misc{flux1,
    author = {{Black Forest Labs}},
    title={FLUX},
    year={2024},
    howpublished={\url{https://github.com/black-forest-labs/flux}},
}

@article{Qwen-Image,
  author       = {{Qwen Team}},
  title        = {Qwen-Image Technical Report},
  journal      = {ArXiv preprint},
  year         = {2025},
}

@article{Wan,
  author       = {{Wan Team}},
  title        = {Wan: Open and Advanced Large-Scale Video Generative Models},
  journal      = {ArXiv preprint},
  year         = {2025},
}

@inproceedings{custom-diffusion,
  author       = {Nupur Kumari and
                  Bingliang Zhang and
                  Richard Zhang and
                  Eli Shechtman and
                  Jun{-}Yan Zhu},
  title        = {Multi-Concept Customization of Text-to-Image Diffusion},
  booktitle    = CVPR,
  year         = {2023},
}

@inproceedings{DreamBooth,
  author       = {Nataniel Ruiz and
                  Yuanzhen Li and
                  Varun Jampani and
                  Yael Pritch and
                  Michael Rubinstein and
                  Kfir Aberman},
  title        = {DreamBooth: Fine Tuning Text-to-Image Diffusion Models for Subject-Driven Generation},
  booktitle    = CVPR,
  year         = {2023},
}

@inproceedings{TextualInversion,
  author       = {Rinon Gal and
                  Yuval Alaluf and
                  Yuval Atzmon and
                  Or Patashnik and
                  Amit Haim Bermano and
                  Gal Chechik and
                  Daniel Cohen{-}Or},
  title        = {An Image is Worth One Word: Personalizing Text-to-Image Generation
                  using Textual Inversion},
  booktitle    = ICLR,
  year         = {2023},
}

@inproceedings{OnlineRL1,
  author       = {Kevin Clark and
                  Paul Vicol and
                  Kevin Swersky and
                  David J. Fleet},
  title        = {Directly Fine-Tuning Diffusion Models on Differentiable Rewards},
  booktitle    = ICLR,
  year         = {2024},
}

@article{OnlineRL2,
  title        = {Aligning Text-to-Image Diffusion Models with Reward Backpropagation},
  journal   = {ArXiv preprint},
  author={Mihir Prabhudesai and Anirudh Goyal and Deepak Pathak and Katerina Fragkiadaki},
  year         = {2023},
}

@article{OnlineRL3,
  author       = {Mihir Prabhudesai and
                  Russell Mendonca and
                  Zheyang Qin and
                  Katerina Fragkiadaki and
                  Deepak Pathak},
  title        = {Video Diffusion Alignment via Reward Gradients},
  journal   = {ArXiv preprint},
  year         = {2024},
}

@article{OnlineRL5,
  author       = {Xiangwei Shen and
                  Zhimin Li and
                  Zhantao Yang and
                  Shiyi Zhang and
                  Yingfang Zhang and
                  Donghao Li and
                  Chunyu Wang and
                  Qinglin Lu and
                  Yansong Tang},
  title        = {Directly Aligning the Full Diffusion Trajectory with Fine-Grained
                  Human Preference},
  journal   = {ArXiv preprint},
  year         = {2025},
}

@article{SRPO,
      title={SRPO: Self-Referential Policy Optimization for Vision-Language-Action Models}, 
      author={Senyu Fei and Siyin Wang and Li Ji and Ao Li and Shiduo Zhang and Liming Liu and Jinlong Hou and Jingjing Gong and Xianzhong Zhao and Xipeng Qiu},
      year={2025},
  journal   = {ArXiv preprint},
}

@inproceedings{policy1,
  author       = {Kevin Black and
                  Michael Janner and
                  Yilun Du and
                  Ilya Kostrikov and
                  Sergey Levine},
  title        = {Training Diffusion Models with Reinforcement Learning},
  booktitle    = ICLR,
  year         = {2024},
}

@inproceedings{policy2,
  author       = {Ying Fan and
                  Kangwook Lee},
  editor       = {Andreas Krause and
                  Emma Brunskill and
                  Kyunghyun Cho and
                  Barbara Engelhardt and
                  Sivan Sabato and
                  Jonathan Scarlett},
  title        = {Optimizing {DDPM} Sampling with Shortcut Fine-Tuning},
  booktitle    = ICML,
  year         = {2023},
}

@inproceedings{policy3,
  author       = {Ying Fan and
                  Olivia Watkins and
                  Yuqing Du and
                  Hao Liu and
                  Moonkyung Ryu and
                  Craig Boutilier and
                  Pieter Abbeel and
                  Mohammad Ghavamzadeh and
                  Kangwook Lee and
                  Kimin Lee},
  title        = {Reinforcement Learning for Fine-tuning Text-to-Image Diffusion Models},
  booktitle    = NeurIPS,
  year         = {2023},
}

@article{trajectory1,
  author       = {Eric Luhman and
                  Troy Luhman},
  title        = {Knowledge Distillation in Iterative Generative Models for Improved
                  Sampling Speed},
  journal   = {ArXiv preprint},
  year         = {2021},
}

@inproceedings{trajectory2,
  author       = {Tim Salimans and
                  Thomas Mensink and
                  Jonathan Heek and
                  Emiel Hoogeboom},
  title        = {Multistep Distillation of Diffusion Models via Moment Matching},
  booktitle    = NeurIPS,
  year         = {2024},
}

@inproceedings{trajectory3,
  author       = {Xingchao Liu and
                  Chengyue Gong and
                  Qiang Liu},
  title        = {Flow Straight and Fast: Learning to Generate and Transfer Data with
                  Rectified Flow},
  booktitle    = ICLR,
  year         = {2023},
}

@inproceedings{trajectory5,
  author       = {Kevin Frans and
                  Danijar Hafner and
                  Sergey Levine and
                  Pieter Abbeel},
  title        = {One Step Diffusion via Shortcut Models},
  booktitle    = ICLR,
  year         = {2025},
}

@article{trajectory6,
  author       = {Hansheng Chen and
                  Kai Zhang and
                  Hao Tan and
                  Leonidas J. Guibas and
                  Gordon Wetzstein and
                  Sai Bi},
  title        = {pi-Flow: Policy-Based Few-Step Generation via Imitation Distillation},
  journal   = {ArXiv preprint},
  year         = {2025},
}

@inproceedings{consistency1,
author = {Song, Yang and Dhariwal, Prafulla and Chen, Mark and Sutskever, Ilya},
title = {Consistency models},
year = {2023},
booktitle = ICML,
}

@inproceedings{consistency2,
  author       = {Jiatao Gu and
                  Chen Wang and
                  Shuangfei Zhai and
                  Yizhe Zhang and
                  Lingjie Liu and
                  Joshua M. Susskind},
  title        = {Data-free Distillation of Diffusion Models with Bootstrapping},
  booktitle    = ICML,
  year         = {2024},
}

@inproceedings{consistency3,
  author       = {Dongjun Kim and
                  Chieh{-}Hsin Lai and
                  Wei{-}Hsiang Liao and
                  Naoki Murata and
                  Yuhta Takida and
                  Toshimitsu Uesaka and
                  Yutong He and
                  Yuki Mitsufuji and
                  Stefano Ermon},
  title        = {Consistency Trajectory Models: Learning Probability Flow {ODE} Trajectory
                  of Diffusion},
  booktitle    = ICLR,
  year         = {2024},
}

@inproceedings{distribution1,
  author       = {Tianwei Yin and
                  Micha{\"{e}}l Gharbi and
                  Richard Zhang and
                  Eli Shechtman and
                  Fr{\'{e}}do Durand and
                  William T. Freeman and
                  Taesung Park},
  title        = {One-Step Diffusion with Distribution Matching Distillation},
  booktitle    = CVPR,
  year         = {2024},
}

@inproceedings{distribution2,
  author       = {Tianwei Yin and
                  Micha{\"{e}}l Gharbi and
                  Taesung Park and
                  Richard Zhang and
                  Eli Shechtman and
                  Fr{\'{e}}do Durand and
                  Bill Freeman},
  title        = {Improved Distribution Matching Distillation for Fast Image Synthesis},
  booktitle    = NeurIPS,
  year         = {2024},
}

@inproceedings{distribution3,
  author       = {Axel Sauer and
                  Dominik Lorenz and
                  Andreas Blattmann and
                  Robin Rombach},
  title        = {Adversarial Diffusion Distillation},
  booktitle    = ECCV,
  year         = {2024},
}

@inproceedings{LoRA,
  author       = {Edward J. Hu and
                  Yelong Shen and
                  Phillip Wallis and
                  Zeyuan Allen{-}Zhu and
                  Yuanzhi Li and
                  Shean Wang and
                  Lu Wang and
                  Weizhu Chen},
  title        = {LoRA: Low-Rank Adaptation of Large Language Models},
  booktitle    = ICLR,
  year         = {2022},
}

@article{T-LoRA,
  author       = {Vera Soboleva and
                  Aibek Alanov and
                  Andrey Kuznetsov and
                  Konstantin Sobolev},
  title        = {T-LoRA: Single Image Diffusion Model Customization Without Overfitting},
  journal   = {ArXiv preprint},
  year         = {2025},
}

@inproceedings{Ada-LoRA,
  author       = {Qingru Zhang and
                  Minshuo Chen and
                  Alexander Bukharin and
                  Pengcheng He and
                  Yu Cheng and
                  Weizhu Chen and
                  Tuo Zhao},
  title        = {Adaptive Budget Allocation for Parameter-Efficient Fine-Tuning},
  booktitle    = ICLR,
  year         = {2023},
}

@inproceedings{DoRA,
  author       = {Shih{-}Yang Liu and
                  Chien{-}Yi Wang and
                  Hongxu Yin and
                  Pavlo Molchanov and
                  Yu{-}Chiang Frank Wang and
                  Kwang{-}Ting Cheng and
                  Min{-}Hung Chen},
  title        = {DoRA: Weight-Decomposed Low-Rank Adaptation},
  booktitle    = ICML,
  year         = {2024},
}

@article{schulman2025lora,
  author  = {Schulman, John and {Thinking Machines Lab}},
  title   = {{LoRA} Without Regret},
  year    = {2025},
  url     = {https://thinkingmachines.ai/blog/lora/},
}

@inproceedings{Q-Diffusion,
  author       = {Xiuyu Li and
                  Yijiang Liu and
                  Long Lian and
                  Huanrui Yang and
                  Zhen Dong and
                  Daniel Kang and
                  Shanghang Zhang and
                  Kurt Keutzer},
  title        = {Q-Diffusion: Quantizing Diffusion Models},
  booktitle    = ICCV,
  year         = {2023},
}

@inproceedings{PTQ4DM,
  author       = {Yuzhang Shang and
                  Zhihang Yuan and
                  Bin Xie and
                  Bingzhe Wu and
                  Yan Yan},
  title        = {Post-Training Quantization on Diffusion Models},
  booktitle    = CVPR,
  year         = {2023}
}

@inproceedings{SVDQuant,
  author       = {Muyang Li and
                  Yujun Lin and
                  Zhekai Zhang and
                  Tianle Cai and
                  Xiuyu Li and
                  Junxian Guo and
                  Enze Xie and
                  Chenlin Meng and
                  Jun{-}Yan Zhu and
                  Song Han},
  title        = {SVDQuant: Absorbing Outliers by Low-Rank Component for 4-Bit Diffusion Models},
  booktitle    = ICLR,
  year         = {2025},
}

@article{NVFP4,
  author       = {NVIDIA},
  title        = {Pretraining Large Language Models with {NVFP4}},
  journal   = {ArXiv preprint},
  year         = {2025},
}

@article{QeRL,
  author       = {Wei Huang and
                  Yi Ge and
                  Shuai Yang and
                  Yicheng Xiao and
                  Huizi Mao and
                  Yujun Lin and
                  Hanrong Ye and
                  Sifei Liu and
                  Ka Chun Cheung and
                  Hongxu Yin and
                  Yao Lu and
                  Xiaojuan Qi and
                  Song Han and
                  Yukang Chen},
  title        = {QeRL: Beyond Efficiency - Quantization-enhanced Reinforcement Learning
                  for LLMs},
  journal   = {ArXiv preprint},
  year         = {2025},
}

@inproceedings{QLoRA,
  author       = {Tim Dettmers and
                  Artidoro Pagnoni and
                  Ari Holtzman and
                  Luke Zettlemoyer},
  title        = {QLoRA: Efficient Finetuning of Quantized LLMs},
  booktitle    = NeurIPS,
  year         = {2023},
}

@inproceedings{LoftQ,
  author       = {Yixiao Li and
                  Yifan Yu and
                  Chen Liang and
                  Nikos Karampatziakis and
                  Pengcheng He and
                  Weizhu Chen and
                  Tuo Zhao},
  title        = {LoftQ: LoRA-Fine-Tuning-aware Quantization for Large Language Models},
  booktitle    = ICLR,
  year         = {2024},
}

@inproceedings{QA-LoRA,
  author       = {Yuhui Xu and
                  Lingxi Xie and
                  Xiaotao Gu and
                  Xin Chen and
                  Heng Chang and
                  Hengheng Zhang and
                  Zhengsu Chen and
                  Xiaopeng Zhang and
                  Qi Tian},
  title        = {QA-LoRA: Quantization-Aware Low-Rank Adaptation of Large Language
                  Models},
  booktitle    = ICLR,
  year         = {2024},
}

@misc{insightface_pypi,
  title        = {InsightFace Python Library (Model Zoo: antelopev2)},
  author       = {{InsightFace Contributors}},
  year         = {2022},
  howpublished = {\url{https://pypi.org/project/insightface/0.7/}},
  note         = {Released Nov 28, 2022. Model pack antelopev2: SCRFD-10GF + ResNet100@Glint360K. Accessed 2026-01-21}
}

@inproceedings{deng2019arcface,
  title     = {ArcFace: Additive Angular Margin Loss for Deep Face Recognition},
  author    = {Deng, Jiankang and Guo, Jia and Xue, Niannan and Zafeiriou, Stefanos},
  booktitle = CVPR,
  year      = {2019}
}

@inproceedings{clip,
  author       = {Alec Radford and
                  Jong Wook Kim and
                  Chris Hallacy and
                  Aditya Ramesh and
                  Gabriel Goh and
                  Sandhini Agarwal and
                  Girish Sastry and
                  Amanda Askell and
                  Pamela Mishkin and
                  Jack Clark and
                  Gretchen Krueger and
                  Ilya Sutskever},
  title        = {Learning Transferable Visual Models From Natural Language Supervision},
  booktitle    = ICML,
  year         = {2021},
}

@article{DINOv3,
  author       = {Oriane Sim{\'{e}}oni and
                  Huy V. Vo and
                  Maximilian Seitzer and
                  Federico Baldassarre and
                  Maxime Oquab and
                  Cijo Jose and
                  Vasil Khalidov and
                  Marc Szafraniec and
                  Seung Eun Yi and
                  Micha{\"{e}}l Ramamonjisoa and
                  Francisco Massa and
                  Daniel Haziza and
                  Luca Wehrstedt and
                  Jianyuan Wang and
                  Timoth{\'{e}}e Darcet and
                  Th{\'{e}}o Moutakanni and
                  Leonel Sentana and
                  Claire Roberts and
                  Andrea Vedaldi and
                  Jamie Tolan and
                  John Brandt and
                  Camille Couprie and
                  Julien Mairal and
                  Herv{\'{e}} J{\'{e}}gou and
                  Patrick Labatut and
                  Piotr Bojanowski},
  title        = {DINOv3},
  journal   = {ArXiv preprint},
  year         = {2025},
}

@article{HPDv2,
  author       = {Xiaoshi Wu and
                  Yiming Hao and
                  Keqiang Sun and
                  Yixiong Chen and
                  Feng Zhu and
                  Rui Zhao and
                  Hongsheng Li},
  title        = {Human Preference Score v2: {A} Solid Benchmark for Evaluating Human
                  Preferences of Text-to-Image Synthesis},
  journal   = {ArXiv preprint},
  year         = {2023},
}

@misc{discus0434_aestheticpredictor_v25,
  author       = {discus0434},
  title        = {Aesthetic Predictor v2.5},
  year         = {2024},
  howpublished = {\url{https://github.com/discus0434/aesthetic-predictor-v2-5}},
  note         = {GitHub repository, accessed 2026-01-21}
}

@inproceedings{PickScore,
  author       = {Yuval Kirstain and
                  Adam Polyak and
                  Uriel Singer and
                  Shahbuland Matiana and
                  Joe Penna and
                  Omer Levy},
  title        = {Pick-a-Pic: An Open Dataset of User Preferences for Text-to-Image
                  Generation},
  booktitle    = NeurIPS,
  year         = {2023},
}

@inproceedings{xu2023imagereward,
  title     = {ImageReward: Learning and Evaluating Human Preferences for Text-to-Image Generation},
  author    = {Xu, Jiazheng and Liu, Xiao and Wu, Yuchen and Tong, Yuxuan and Li, Qinkai and Ding, Ming and Tang, Jie and Dong, Yuxiao},
  booktitle = NeurIPS,
  year      = {2023},
}

@misc{nvidia2024blackwell,
  title={NVIDIA Blackwell Architecture Technical Brief},
  author={NVIDIA},
  year={2024},
  url={https://resources.nvidia.com/en-us-blackwell-architecture}
}

@article{mxfp4,
  author       = {Bita Darvish Rouhani and
                  Ritchie Zhao and
                  Ankit More and
                  Mathew Hall and
                  Alireza Khodamoradi and
                  Summer Deng and
                  Dhruv Choudhary and
                  Marius Cornea and
                  Eric Dellinger and
                  Kristof Denolf and
                  Dusan Stosic and
                  Venmugil Elango and
                  Maximilian Golub and
                  Alexander Heinecke and
                  Phil James{-}Roxby and
                  Dharmesh Jani and
                  Gaurav Kolhe and
                  Martin Langhammer and
                  Ada Li and
                  Levi Melnick and
                  Maral Mesmakhosroshahi and
                  Andres Rodriguez and
                  Michael Schulte and
                  Rasoul Shafipour and
                  Lei Shao and
                  Michael Y. Siu and
                  Pradeep Dubey and
                  Paulius Micikevicius and
                  Maxim Naumov and
                  Colin Verilli and
                  Ralph Wittig and
                  Doug Burger and
                  Eric S. Chung},
  title        = {Microscaling Data Formats for Deep Learning},
  journal   = {ArXiv preprint},
  year         = {2023},
}

@inproceedings{yeh2024navigating,
  title={Navigating Text-To-Image Customization: From Ly{CORIS} Fine-Tuning to Model Evaluation},
  author={SHIH-YING YEH and Yu-Guan Hsieh and Zhidong Gao and Bernard B W Yang and Giyeong Oh and Yanmin Gong},
  booktitle=ICLR,
  year={2024},
}

@misc{pyiqa,
  title={{IQA-PyTorch}: PyTorch Toolbox for Image Quality Assessment},
  author={Chaofeng Chen and Jiadi Mo},
  year={2022},
  howpublished = "[Online]. Available: \url{https://github.com/chaofengc/IQA-PyTorch}"
}

@inproceedings{BLIP,
  author       = {Junnan Li and
                  Dongxu Li and
                  Caiming Xiong and
                  Steven C. H. Hoi},
  editor       = {Kamalika Chaudhuri and
                  Stefanie Jegelka and
                  Le Song and
                  Csaba Szepesv{\'{a}}ri and
                  Gang Niu and
                  Sivan Sabato},
  title        = {{BLIP:} Bootstrapping Language-Image Pre-training for Unified Vision-Language
                  Understanding and Generation},
  booktitle    = ICML,
  year         = {2022},
}

@inproceedings{aghajanyan2020intrinsic,
  author       = {Armen Aghajanyan and
                  Sonal Gupta and
                  Luke Zettlemoyer},
  title        = {Intrinsic Dimensionality Explains the Effectiveness of Language Model
                  Fine-Tuning},
  booktitle    =  {International Joint Conference on Natural
                  Language Processing},
  year         = {2021},
}

@inproceedings{houlsby2019parameter,
  author       = {Neil Houlsby and
                  Andrei Giurgiu and
                  Stanislaw Jastrzebski and
                  Bruna Morrone and
                  Quentin de Laroussilhe and
                  Andrea Gesmundo and
                  Mona Attariyan and
                  Sylvain Gelly},
  title        = {Parameter-Efficient Transfer Learning for {NLP}},
  booktitle    = ICML,
  year         = {2019},
}

@inproceedings{SparseLoRA,
  author       = {Samir Khaki and
                  Xiuyu Li and
                  Junxian Guo and
                  Ligeng Zhu and
                  Konstantinos N. Plataniotis and
                  Amir Yazdanbakhsh and
                  Kurt Keutzer and
                  Song Han and
                  Zhijian Liu},
  title        = {SparseLoRA: Accelerating {LLM} Fine-Tuning with Contextual Sparsity},
  booktitle    = ICML,
  year         = {2025},
}
\bibliographystyle{icml2026}

\newpage
\appendix
\onecolumn
\section{Detailed Experimental Settings}

\subsection{Customization Evaluation Pipeline}

To ensure a comprehensive evaluation, we select five distinct concepts within each category, each exhibiting substantial visual variance.
Following standard few-shot customization protocols, we curate five independent prompt--image pairs per concept.
This high-variance construction is designed to probe the model’s generalization capability and training stability across a diverse set of visual concepts.

During evaluation, we construct five sets of diverse test prompts for each concept.
As discussed in Section~\ref{sec:exp}, customization quality is quantified using task-specific feature-space metrics.
Specifically, for \emph{Human Identity}, we employ AntelopeV2 to extract facial embeddings and compute cosine similarity with the training samples~\cite{insightface_pypi,deng2019arcface};
for \emph{Artistic Style}, we use CLIP to measure stylistic alignment between generated and reference images~\cite{clip};
and for \emph{General Subject}, we adopt DINOv3 to extract fine-grained visual features for evaluating structural and textural consistency~\cite{DINOv3}.
In addition, we assess image quality using PyIQA~\cite{pyiqa},
measure diversity by averaging pairwise DINOv3 embedding distances among generated images,
and evaluate prompt-following behavior using BLIP~\cite{BLIP}.

\subsection{Customization Dataset}

\begin{figure}[h]
    \centering
    \begin{minipage}{0.32\linewidth}
        \centering
        \includegraphics[width=\linewidth]{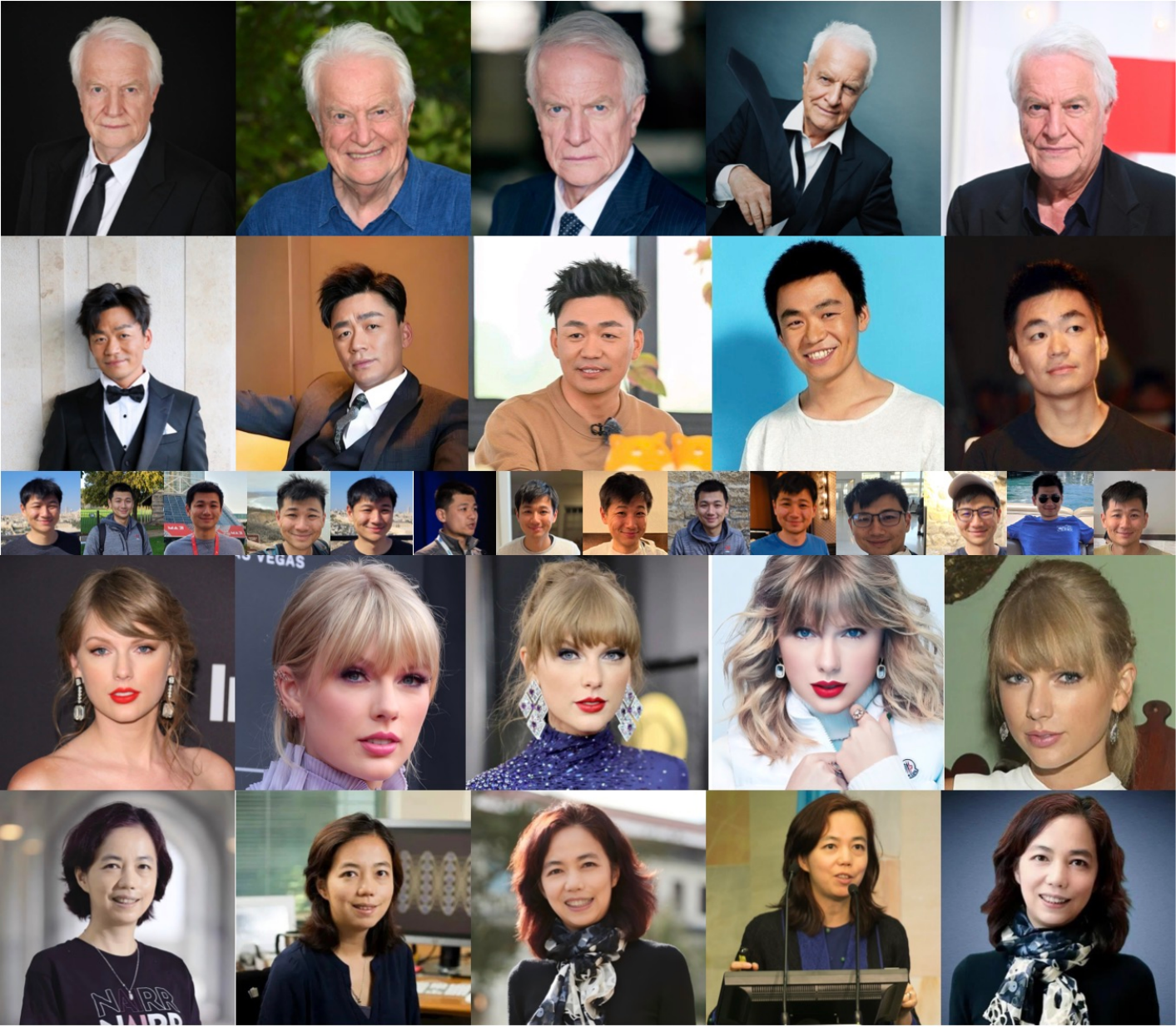}
        \caption*{(a) Human Identity}
    \end{minipage}
    \hfill
    \begin{minipage}{0.32\linewidth}
        \centering
        \includegraphics[width=\linewidth]{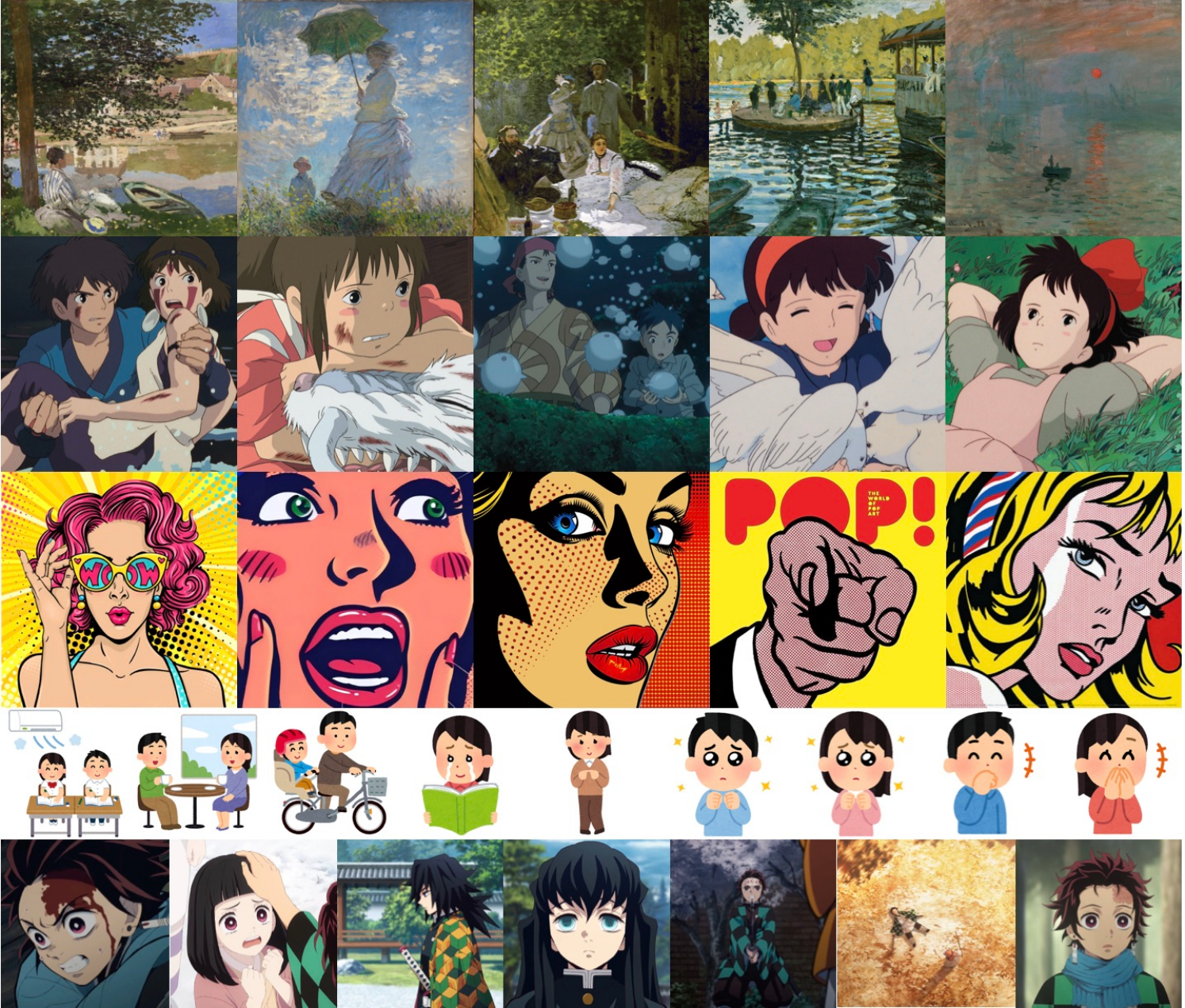}
        \caption*{(b) Artistic Style}
    \end{minipage}
    \hfill
    \begin{minipage}{0.32\linewidth}
        \centering
        \includegraphics[width=\linewidth]{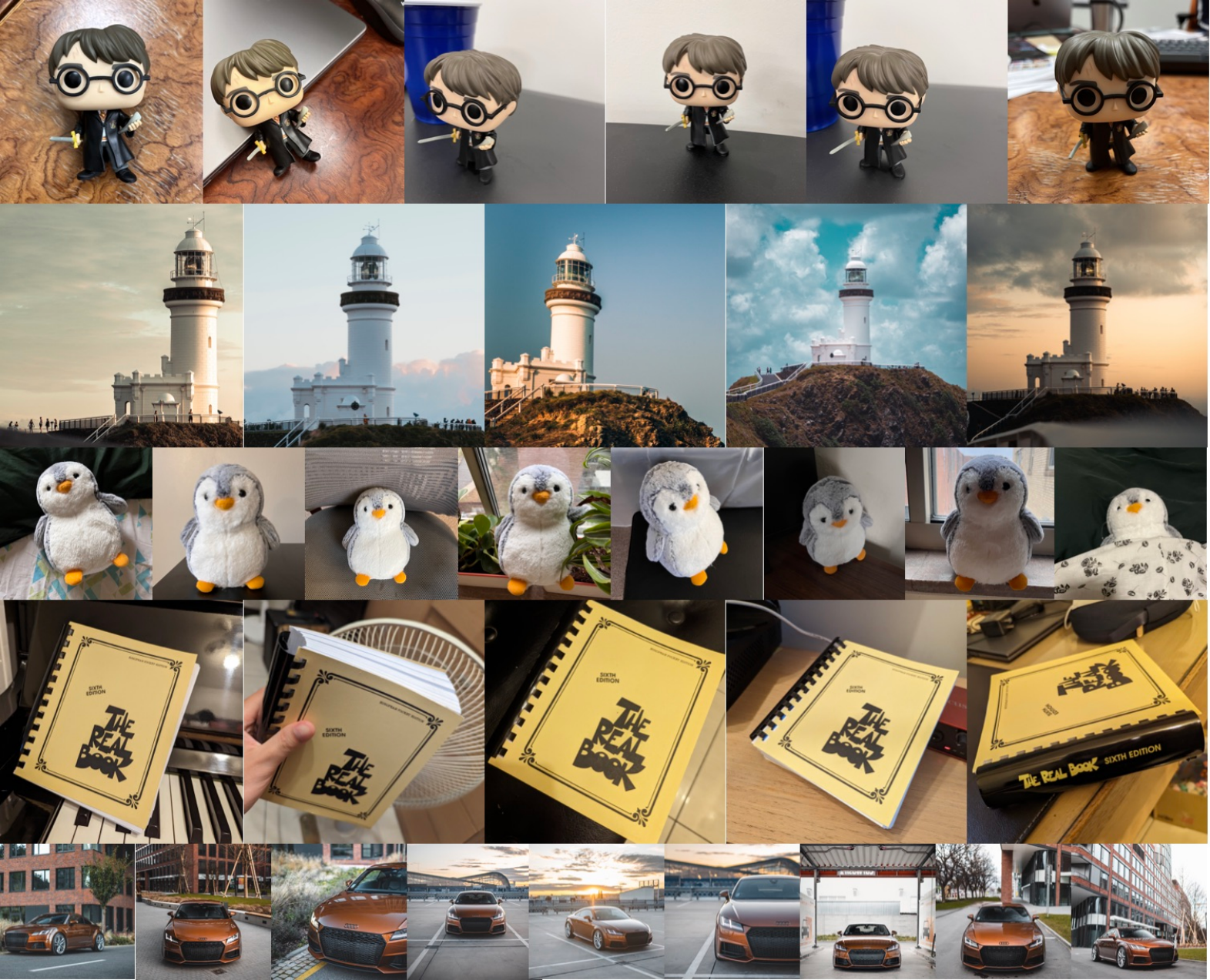}
        \caption*{(c) General Subject}
    \end{minipage}

    \caption{
     Hybrid customization dataset covering three categories: Human Identity, Artistic
Style, and General Subject.
    }
    \label{fig:customization_dataset}
\end{figure}

Figure~\ref{fig:customization_dataset} shows the complete dataset used on customization tasks.
The dataset consists of three categories:
Human Identity, which focuses on preserving personal identity while allowing variations in appearance or attributes.
Artistic Style, which targets adapting artistic or visual styles while keeping the underlying content unchanged.
General Subject, which emphasizes generating a specific subject consistently across different contexts.

\subsection{Customization Metrics}
We evaluate customization behavior using two complementary metrics:
\emph{diversity} and \emph{Prompt Following}.
Diversity is measured using DINO image embeddings to capture visual variation,
while Prompt Following is measured using BLIP to assess prompt-image alignment.

\paragraph{Diversity.}
Diversity measures the visual variation among generated images for the same customization task.
Given a set of generated images $\mathcal{I}=\{i_1,\ldots,i_n\}$,
we extract DINO image embeddings $d(i)$ and compute the average pairwise cosine similarity.
The diversity score is defined as
\[
D = 1 - \frac{1}{\binom{n}{2}} \sum_{j<k}
\mathrm{cos}\big(d(i_j), d(i_k)\big),
\]
where larger values indicate higher visual diversity.

\paragraph{Prompt Following.}
Prompt Following evaluates how well generated images follow the input prompts.
Given paired prompts and images $\{(p_i,i_i)\}_{i=1}^{n}$,
we compute cosine similarity between BLIP text and image embeddings.
The instruction fidelity score is defined as
\[
F = \frac{1}{n} \sum_{i=1}^{n}
\mathrm{cos}\big(e(p_i), e(i_i)\big),
\]
where $e(\cdot)$ denotes BLIP embeddings and larger values indicate better
prompt-image alignment.

\section{Generalization}
\label{sec:generalization}

Although our main experiments focus on large-scale DiT backbones, i.e., FLUX.1-dev and Qwen-Image, we further evaluate whether FourTune generalizes beyond the main experimental setup.
We consider two complementary settings: transferring to a conventional latent diffusion backbone, SDXL, and running FourTune with an INT4 implementation on an NVIDIA RTX 4090 GPU.

\begin{table}[!ht]
\centering
\caption{
Generalization to SDXL on identity customization.
Our W4A4G4 method maintains generation quality comparable to the 16-bit LoRA baseline.
}
\label{tab:generalization_sdxl}
\small
\setlength{\tabcolsep}{5pt}
\renewcommand{\arraystretch}{1.35}

\begin{tabular}{cccccc}
\toprule
Method & Precision & Similarity & \makecell{Image\\Quality} & Diversity & \makecell{Prompt\\Following} \\
\midrule
BF16 LoRA & W16A16G16 & \textbf{0.454} & 11.17 & 0.658 & 0.935 \\
\rowcolor{ourscolor!40}
\textbf{Ours} & W4A4G4 & 0.453 & \textbf{15.33} & \textbf{0.664} & \textbf{0.976} \\
\bottomrule
\end{tabular}
\end{table}

As shown in Table~\ref{tab:generalization_sdxl}, FourTune transfers well to SDXL under the same W4A4G4 training pipeline.
Its overall customization quality remains competitive with the BF16 LoRA baseline, suggesting that the proposed 4-bit training scheme is not restricted to DiT architectures.

\begin{table}[!ht]
\centering
\caption{
Generalization across quantization formats and hardware/software stacks on FLUX.1-dev identity customization.
The INT4 variant is evaluated on an RTX 4090 GPU and achieves a $2.6\times$ speedup over QLoRA.
}
\label{tab:generalization_int4}
\small
\setlength{\tabcolsep}{5pt}
\renewcommand{\arraystretch}{1.35}

\begin{tabular}{cccccc}
\toprule
Method & Precision & Similarity & \makecell{Image\\Quality} & Diversity & \makecell{Prompt\\Following} \\
\midrule
BF16 LoRA & W16A16G16 & 0.771 & 33.49 & \textbf{0.577} & 0.941 \\
\rowcolor{ourscolor!40}
\textbf{Ours} (NVFP4) & W4A4G4 & \textbf{0.783} & \textbf{34.77} & 0.570 & 0.913 \\
\rowcolor{ourscolor!40}
\textbf{Ours} (INT4) & W4A4G4 & 0.777 & 31.64 & 0.545 & \textbf{0.962} \\
\bottomrule
\end{tabular}
\end{table}

FourTune also remains effective under a separate INT4 runtime on the Ada architecture.
As shown in Table~\ref{tab:generalization_int4}, both NVFP4 and INT4 variants achieve competitive generation quality compared with BF16 LoRA, while the INT4 implementation delivers a $2.6\times$ speedup over QLoRA.
These results suggest that FourTune generalizes across model architectures, low-bit formats, and hardware/software stacks.

\end{document}